%% file: sn-article.tex
\documentclass[pdflatex,sn-basic]{sn-jnl}% Basic Springer Nature Reference Style/Chemistry Reference Style
\input{mathinput.tex}

\usepackage{subcaption}
\usepackage{amsmath,amssymb,amsfonts}%
\usepackage{xcolor}%
\usepackage{booktabs}%
\definecolor{vertFonce}{rgb}{0, 0.6, 0}

\begin{document}

\title[Deep Unsupervised Domain Adaptation for Time Series Classification: a Benchmark]{Deep Unsupervised Domain Adaptation for Time Series Classification: a Benchmark}

\author{\fnm{Hassan} \sur{Ismail Fawaz}}
\author{\fnm{Ganesh} \sur{Del Grosso}}
\author{\fnm{Tanguy} \sur{Kerdoncuff}}
\author{\fnm{Aurélie} \sur{Boisbunon}}
\author{\fnm{Illyyne} \sur{Saffar}}

\affil{\orgdiv{AI Research \& Systems}, \orgname{Ericsson Research}, \orgaddress{\city{Paris}, \country{France}}. \\\{first.last-name\}@ericsson.com}


\abstract{
Unsupervised Domain Adaptation (UDA) aims to harness labeled source data to train models for unlabeled target data. 
While research in UDA is extensive in domains like computer vision and natural language processing, it remains under-explored for time series data despite its widespread real-world applications. 
Our paper addresses this gap by introducing a comprehensive benchmark for evaluating UDA techniques for time series classification, with a focus on deep learning methods. 
We provide a fair and standardized UDA method assessment with state of the art neural network backbones (e.g. Inception) for time series data, as well as seven extra datasets in addition to the already established ones.
This benchmark offers insights into the strengths and limitations of the evaluated approaches while preserving the unsupervised nature of DA, making it directly applicable to real data mining problems. 
It serves as a beneficial resource for researchers and practitioners, fostering innovation in this critical field. In order to ensure reproducibility, %we publish our full results and code here: \url{https://github.com/EricssonResearch/UDA-4-TSC}.
the code and results have been open-sourced\footnote{\url{https://github.com/EricssonResearch/UDA-4-TSC}}.
}

\keywords{benchmark, unsupervised domain adaptation, deep learning, time series classification}



\maketitle
\section*{Acknowledgments}

First, We would like to express our sincere gratitude to all individuals managing the AILAB Server within the Ericsson Research department. Their support and assistance have been invaluable for the completion of these research experiments.

A special acknowledgment is extended to Laetitia Chapel, Charlotte Pelletier and Romain Tavenard from the IRISA lab (Institute for Research in Computer Science and Random Systems), for their collaborative spirit and the resources they shared with us. Furthermore, we deeply appreciate their critical review and constructive feedback on earlier drafts of our work, which significantly enhanced the clarity and coherence of our findings.

Finally, we would also like to express our sincere gratitude to the researchers who collected, cleaned, and curated the datasets used in this benchmark. 
Their invaluable contribution forms the foundation of our study, and we appreciate their efforts in providing these datasets for free.

\section{Introduction}\label{sec1}

The realm of time series data mining has witnessed a remarkable surge in recent years and has become central in diverse cognitive tasks, 
driven by the omnipresence of time series data across multiple domains and the advent of the Internet of Things (IoT).
The manipulation and analysis of temporal sequences, characterized by an inherent ordering, 
have become integral components of data science practices spanning fields such as medicine, telecommunications, remote sensing, and human activity recognition~\citep{sanz2022exploring,kamalov2021pazoe}. 
Within this domain, Time Series Classification (TSC) entails the labeling of a time series based on a set of labeled training examples~\citep{ismail2019deep}, 
to identify, such as the type of crop in a field or specific human actions.

Recently, the literature in TSC showed substantial advancements in various applications~\citep{ruiz2021great}. 
However, despite this success, state-of-the-art approaches fail to address the discrepancy arising between the data observed in the deployment environment of a machine learning model and the data employed during its training phase. 
Such a disparity occurs as a result of the unique characteristics inherent to the new operational context: for instance, variations in genomic profiles or the medical condition of a patient~\citep{ptbXLecg}, fluctuations in the density of devices or shifts in patterns of communication data utilization~\citep{park2022predicting}, changes in weather conditions and soil properties within remote sensing applications~\citep{nyborg2022timematch}.
An analogous situation occurs when models are trained using synthetic data then later used for real-world data analysis~\citep{webb2016characterizing}. 
Consequently, it becomes imperative to adapt the model to accommodate this data distribution shift, which constitutes the fundamental objective of domain adaptation. 
This challenge assumes an even greater degree of complexity when labels are entirely absent in the new domain, thereby giving rise to the subfield of unsupervised domain adaptation.

Domain Adaptation (DA) and Unsupervised Domain Adaptation (UDA) have garnered substantial attention in the fields of Natural Language Processing (NLP) and Computer Vision (CV)~\citep{ruder2019transfer, li2020transfer, xu2022video}. 
These domains have witnessed the emergence of numerous methodological approaches to address data distribution disparities.
One commonly employed strategy involves the transfer of specific layers from Neural Networks (NN), followed by the freezing of these layers during the forward pass~\citep{fawaz2018transfer}. 
In the context of data shift, particularly covariate shift, more intricate algorithms have been developed.
These include classification algorithms based on adversarial training~\citep{ganin2016domain,long2018conditional}, contrastive learning~\citep{kang2019contrastive} or on the discrepancy between domains~\citep{OTDA, damodaran2018deepjdot}. 
In contrast, other works concentrate on DA regression tasks, for instance \cite{wei2022adaptive} propose to utilize a domain-aware kernel for comparing samples.
To assess the efficacy of these methodologies, benchmarking initiatives have been established~\citep{zhao2020review,ringwald2021adaptiope}.
In~\cite{ringwald2021adaptiope} authors conducted a comparative evaluation of these approaches using image datasets, revealing notable challenges related to image noise and annotation errors while demonstrating significant improvements upon addressing them.
In~\cite{zhao2020review} researchers concentrated predominantly on evaluating deep unsupervised methods for vision tasks within their benchmarking framework.

\begin{figure}
    \centering
    \includegraphics[width=0.85\textwidth]{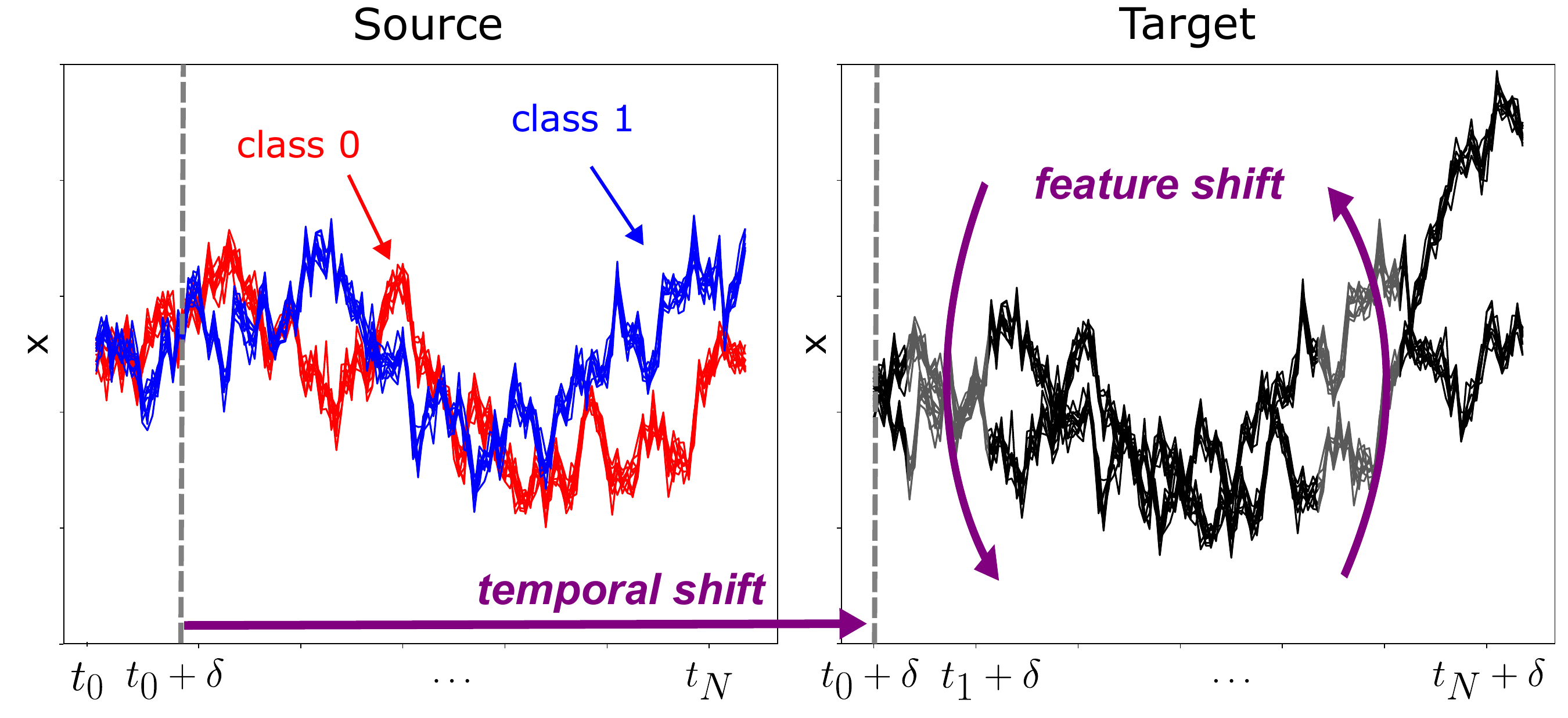}
    \caption{Illustration of the types of shift in unsupervised domain adaptation for time series data: temporal shift occur when the shift is captured on the time axis, where the start of the measurements do not necessarily match the same moment of the time series, while feature shift occur when the shift is captured on the feature space. This example displays the case where both types of shift are present in target data.}
    \label{fig:shift_tS}
\end{figure}

In contrast to the well-established benchmarking frameworks in computer vision and NLP, the time series data mining community currently lacks a comprehensive overview of the existing techniques and datasets available for benchmarking UDA in the context of TSC, as highlighted in \cite{sasa}.
UDA for time series presents unique challenges due to the inherent characteristics of time series data. Unlike images, time series data involves temporal ordering, trends, seasonalities, and varying frequencies, all of which influence analysis and performance. These factors contribute to the increased complexity of UDA for time series. Fig. \ref{fig:shift_tS} illustrates some challenges of UDA in time series, highlighting that shifts can occur temporally, in the feature space, or both simultaneously.
While an initial work in this direction has been made by~\cite{ragab2023adatime}, it remains constrained in terms of data diversity, dataset quantity, and statistical analysis for assessing the performance of time series classifiers across domains. 
The benchmark we propose compares the performance of 9 UDA algorithms integrated with cutting-edge backbone architectures on a set of 12 datasets, including 7 coming from multivariate time series applications that we tailored to domain adaptation to diversify domain contexts.
Additionally, we delve into the choice of hyperparameter tuning criteria, a crucial aspect in UDA given the absence of labeled data in the target domain. We also provide insights on how the performance of UDA approaches is impacted by data characteristics, such as shift between source and target, number of classes, and class imbalance.
This enhanced analysis establishes valuable insights into the adaptability and robustness of UDA algorithms across different time series scenarios, fostering a more nuanced and informed understanding of their capabilities.

%\begin{figure}[!ht]
%  \centering
%\begin{subfigure}{.8\textwidth}
%    \centering
%    \includegraphics[width=0.99\textwidth]{fig/temporalshift.png}
%    \caption{Temporal Shift}
%    \label{tempShift}
%\end{subfigure}
%
%\begin{subfigure}{.8\textwidth}
%    \centering
%    \includegraphics[width=0.99\textwidth]{fig/featureshift.png}
%    \caption{Feature Shift}
%    \label{featShift}
%\end{subfigure}
%
%\begin{subfigure}{.8\textwidth}
%    \centering
%   \includegraphics[width=0.99\textwidth]{fig/dualshift.png}
%    \caption{Dual Shift}
%    \label{dualShift}
%\end{subfigure}
%
%\caption{Type of shifts in time series data: (a) Temporal shift: where the shift is captured on the time axis, (b) Feature shift: where the shift is captured on the feature space and (c) Dual shift: where the shift is both on the time axis and on the feature space at the same time.%\textcolor{red}{I think we can 1) remove axis value (we don’t care about the scale) but we can increase the ticks size 2) increase slightly the size of the text value,time,source,target. Maybe we can remove the dual shift so that we can increase the size of the figure ? Better also to have them in pdf and split. 3) I (Hassan) think we can have the figures on two rows, each row has three figures, top row shows source, bottom row shows target, that way we can have larger figure.}
%% not the biggest priority, will be updated later
%}
%\label{fig:shift_tS}
%\end{figure}

\section{Unsupervised Domain Adaptation Background}

This section starts by defining the problem statement followed by a list of existing approaches to tackle the UDA task. 
We then introduce three main popular techniques for tuning the hyperparameters under the DA regime. 
Finally, we explore the open question of estimating the shift between the distribution of two TSC datasets by introducing a novel shift proxy that leverages a strong neural network time series classifier. %\textcolor{red}{An illustration of UDA task for TCS with different shift is illustrated on Fig.~\ref{fig:shift_tS}.}

\subsection{Notations and Problem Statement}

Let $D_s=\{(\rmX_i^s, \rvy_i^s)\}^{n_s}_{i=1}$ be a source domain consisting of $n_s$ labeled time series realizations of variables $(X, y)$, sampled from a distribution $p_s$, with $\rmX_i^s \in \mathcal{X}$ the time series data and $\rvy_i^s \in \mathcal{Y}$ the corresponding labels, and let $D_t=\{\rmX_j^t\}^{n_t}_{j=1}$ represent a target domain solely consisting of $n_t$ unlabeled time series, sampled from another distribution $p_t$.
The objective of UDA is to learn a classifier capable of accurately estimating the label of a time series in the target domain by making use of the knowledge learned from the labeled time series in source.
In mathematical terms, a good classifier $f$ should have a low target risk:
\begin{equation}
    \mathcal{R}_t(f) =\mathbb{E}_{(X,y)\sim p_t}\left[  \Ls (f(X),y)\right]
    \label{eq:target-risk}
\end{equation} 
for a given loss $\Ls$.
As some datasets contain multiple domains (\textit{e.g.}, patients or geographical areas) each adaptation from one domain to another is referred to as a \textit{scenario}.

In the majority of the DA literature, the theoretical findings~\citep{ben2006analysis} as well as the underlying motivation of the algorithms~\citep{ganin2016domain} presume that the shift between the two domains adheres to the covariate shift assumption~\citep{farahani2021brief}.
The latter posits that the data distribution differs between source and target domains, while the conditional distribution of labels, given the input data, remains unchanged, i.e., %$p_s(\rmX)\neq p_t(\rmX)$ but $p_s(\rvy|\rmX)=p_t(\rvy|\rmX)$ for any $\rmX\in\mathcal{X}$ and $\rvy\in\mathcal{Y}$.
$p_s(X)\neq p_t(X)$ but $p_s(y|X)=p_t(y|X)$ for any $X\in\mathcal{X}$ and $y\in\mathcal{Y}$.

\subsection{UDA Algorithms for time series classification}
\label{sec:algo}

The focus of this work is on deep UDA algorithms for TSC, which rely on two main components: a backbone encoding the input into a domain-invariant latent space, and a classifier. 
Labeled data from the source domain are used to train the classifier for the main classification task, while unlabeled data from the target domain are used to adapt the model to the target domain.
Table \ref{tab:Algos} summarizes the algorithms considered in this work. 
Furthermore, in the following subsections, we describe these algorithms in more details.

\begin{table*}[t]
    \centering
    \caption{Deep UDA algorithms for TSC (top) and baseline
    (bottom). They consist of a backbone, a classifier and other specific modules. $\mathcal{L}_{C}$, $\mathcal{L}_{A}$, $\mathcal{L}_\mathrm{VRNN}$, $\mathcal{L}_{R}$, $\mathcal{L}_{\mathrm{Contrastive}}$, $H$,  and $\mathcal{L}_{\mathrm{Sinkhorn}}$ denote the classification, adversarial, VRRN, reconstrution and contrastive loss functions, the entropy and the Sinkhorn divergence respectively.}
    \label{tab:Algos}
    \small
    {
    \begin{tabular}{l c c c }
    \toprule
        Algorithms & Backbone & Other Modules & Loss function\\
        \midrule
        %\hline \hline
        VRADA & VRNN & Discriminator & $\mathcal{L}_{C} + \mathcal{L}_{A} + \mathcal{L}_\mathrm{VRNN}$\\
        CoDATS & 1D CNN & Discriminator & $\mathcal{L}_{C} + \mathcal{L}_{A}$\\
        InceptionDANN & Inception & Discriminator & $\mathcal{L}_{C} + \mathcal{L}_{A}$\\
        InceptionCDAN & Inception & Discriminator, Multilinear Map & $\mathcal{L}_{C} + \mathcal{L}_{A}$\\
        CoTMix & 1D CNN & Temporal Mixup & $\mathcal{L}_{C} + \mathcal{L}_{\mathrm{Contrastive}} + H$\\
        InceptionMix & Inception & Temporal Mixup & $\mathcal{L}_{C} + \mathcal{L}_{\mathrm{Contrastive}} + H$ \\
        Raincoat & 1D CNN & Frequency encoder, Decoder & $\mathcal{L}_{C} + \mathcal{L}_{R} + \mathcal{L}_{\mathrm{Sinkhorn}}$ \\
        InceptionRain & Inception & Frequency encoder, Decoder & $\mathcal{L}_{C} + \mathcal{L}_{R} + \mathcal{L}_{\mathrm{Sinkhorn}}$\\ 
        \midrule
        InceptionTime & Inception & - & $\mathcal{L}_{C}$ \\
        %OTDA & - & Transport map & - \\
        \bottomrule
    \end{tabular}
    }
\end{table*}

\subsubsection{Baseline} 
%\tanguy{As previously presented, $\mathcal{X}$ is the input space and $\mathcal{Y}$ a label vector space}
%Let $\mathcal{X}$ be the input space. Let $\mathcal{Y}$ be the label space,a vector space where the labels represented as one-hot vectors and the predictions represented as categorical probabilities live. $D_s$ is the source domain consisting of $n_s$ labeled time series. For a sample $(\rmX^s,\rvy^s)\in D_s$ let $\rmX^s\in\mathcal{X}$ represent the time series data and $\rvy^s\in\mathcal{Y}$ represent the corresponding label. $D_t\subset\mathcal{X}$ is the target domain consisting solely of $n_t$ unlabeled time series. $\mathcal{Z}$ is the latent space.
The baseline is composed of a neural network backbone $E:\mathcal{X}\rightarrow\mathcal{Z}$ with $\mathcal{Z}$ a latent space.
In the case of time
series data, the backbone can, for instance, be a Recurrent Neural Network~\citep{yu2019review} or a Convolutional Neural Network~\citep{li2021survey}. 
%Every backbone is associated with the same classifier $C:\mathcal{Z}\rightarrow\mathcal{Y}$, mapping the latent space representation of the input into a categorical probability.
The backbone is then associated with a classifier $C:\mathcal{Z}\rightarrow\mathcal{Y}$, mapping the latent space representation of the input into a categorical probability.

The NN will be optimized by minimizing the cross-entropy loss function $\mathcal{L}:\mathcal{Y}\times\mathcal{Y}\rightarrow\RR$, which is averaged over all source samples to get the classification loss $\mathcal{L}_{C}$,
%\begin{equation*}
%    \mathcal{L}_{C}(E,C) = \frac{1}{n_s}\sum_{i=1}^{n_s}\mathcal{L}\left(C\left(E(\rmX^s_i)\right),\rvy^s_i\right).
%\end{equation*}
\begin{equation}
    \mathcal{L}_{C}(E,C) = \mathbb{E}_{(X, y)\sim p_s}\left[\mathcal{L}\left(C\left(E(X)\right),y\right)\right].
\end{equation}

We use InceptionTime~\citep{InceptionTime} as a baseline with no adaptation to compare with UDA approaches and highlight the benefits of applying domain adaptation.
%This state-of-the-art approach for time series classification is trained with the source domain data, and the target domain data is used only for hyperparameter tuning if needed. 
Additionally, since it has been by shown \cite{ruiz2021great} to be the current best deep architecture for TSC, we propose to use its backbone, referred to as the ``Inception" backbone, for other domain adaptation algorithms, thus allowing the comparison of different UDA algorithms independently of their backbones.

\subsubsection{Adversarial domain adaptation} 

Inspired by Generative Adversarial Networks~\citep{GANS}, \cite{ganin2016domain} propose a Domain-Adversarial training of Neural Network (DANN) to force the backbone to produce a common domain invariant latent representation for both source and target.
For this purpose, a discriminator
%$D:\mathcal{Z}\rightarrow\{[0,1], [0,1]\}$
$D:\mathcal{Z}\rightarrow[0,1]$
is introduced with the task of separating source and target domain samples.
The discriminator learns how to separate source domain samples from target domain samples and the backbone learns how to fool the discriminator using the following adversarial loss based on cross-entropy,
%\begin{eqnarray*}
%    \mathcal{L}_{A}(E,D) = \frac{1}{n_s}\sum_{i=1}^{n_s}\mathcal{L}\left(D\left(E(\rmX^s_i)\right),(0,1)\right)
%    + \frac{1}{n_t}\sum_{i=1}^{n_t}\mathcal{L}\left(D\left(E(\rmX^t_i)\right),(1,0)\right),
    %\label{eq:adv_loss}
%\end{eqnarray*}
\begin{eqnarray}
    \mathcal{L}_{A}(E,D) = -\mathbb{E}_{(X,y)\sim p_s}\left[\log\left(D\left(E\left(X\right)\right)\right)\right]
    - \mathbb{E}_{(X,y)\sim p_t}\left[\log\left(1 - D\left(E\left(X\right)\right)\right)\right].
    %\label{eq:adv_loss}
\end{eqnarray}
The total loss for the adversarial domain adaptation framework is
\begin{equation}
    \mathcal{L}_{\mathrm{total}}(E,C,D) = \mathcal{L}_{C}(E,C) -\lambda\mathcal{L}_{A}(E,D)\;,
    \label{eq:adv_loss_total}
\end{equation}
where $\lambda$ is an hyperparameter that balances the adversarial and classification losses.
The key part of adversarial learning is that, during training, the total loss in (\ref{eq:adv_loss_total}) is minimized with respect to the weights of classifier $C$ and encoder $E$ and maximized with respect to the weights of discriminator $D$.
Because DANN was originally not designed for TSC, we will use instead a 1-dimensional fully convolutional neural network (1D CNN) for the backbone. This is motivated by the paper Convolutional deep Domain Adaptation model for Time Series data~\citep[CoDATS, ][]{CoDATS} that applies DANN to time series.
Additionaly, we also use Inception backbone (InceptionDANN) previously presented.

Variational Recurrent Adversarial Deep Domain Adaptation~\citep[VRADA, ][]{VRADA} is also based on the adversarial approach but uses additionaly a Variational Recurrent Neural Network~\citep[VRNN, ][]{VRNNs} as a backbone in order to achieve the variability observed in highly structured time series data.
The learning objective of a VRNN ($\mathcal{L}_\mathrm{VRNN}$) is to minimize the distance in distribution between the inference model generated by the VRNN and a prior and to minimize the log-likelihood of the reconstructed input.
%The VRNN loss is a function of the VRNN itself and we use $\mathcal{L}_\mathrm{VRNN}$ to refer to it.
The total loss for VRADA is
\begin{eqnarray}
    \mathcal{L}_{\mathrm{total}}(E,C,D) = \mathcal{L}_\mathrm{VRNN}(E) + \mathcal{L}_{C}(E,C) -\lambda\mathcal{L}_{A}(E,D),
    \label{eq:vrada_loss}
\end{eqnarray}
with the same minimization with respect to $C$ and $E$ and maximization with respect to $D$ as in equation~\ref{eq:adv_loss_total}.
%This objective is minimized with respect to the classifier $C$ and encoder $E$ and maximized with respect to the discriminator $D$.

The Conditional Domain Adversarial Network (CDAN) algorithm~\citep{long2018conditional}, based on DANN, also tries to enforce domain adaptation through adversarial learning. But it requires additionally a multilinear map $T:\mathcal{Z}\times\mathcal{Y}\rightarrow\mathcal{Z}'$, the simplest implementation computes the outer product between the latent space representation and the soft probabilities of the classifier, thus conditioning the input of the discriminator on the predicted class.
With the new input space of $D$ being $\mathcal{Z}'$, the adversarial loss reads
%\begin{eqnarray*}
%    &\mathcal{L}_{A}(E,C,D) &\\
%    &=& \frac{1}{n_s}\sum_{i=1}^{n_s}\mathcal{L}\left(D_{T,C}(\rmX^s), \rvy^s_\mathrm{domain}\right) \\
%    &&+ \frac{1}{n_t}\sum_{i=1}^{n_t}\mathcal{L}\left(D_{T,C}(\rmX^t) ,\rvy^t_\mathrm{domain}\right)\;,
    %\label{eq:adv_loss_multi}
%\end{eqnarray*}
\begin{eqnarray}
    \mathcal{L}_{A}(E,C,D) = -\mathbb{E}_{(X,y)\sim p_s}\left[\log\left(D_{T,C}\left(X\right)\right)\right]
    - \mathbb{E}_{(X,y)\sim p_t}\left[\log\left(1 - D_{T,C}\left(X\right)\right)\right]
    \label{eq:adv_loss_multi2}
\end{eqnarray}
where $D_{T,C}(\rmX) = D\left(T\left(E(\rmX),C(E(\rmX))\right)\right)$.
Both the randomized multilinear map and entropy conditioning \citep{long2018conditional} are implemented and the different options are simply considered as a hyperparameter.
Again, as the original backbone used was not designed for time series, we use Inception (InceptionCDAN).

%In summary, this technique uses a discriminator for distinguishing between source and target samples, while the backbone learns simultaneously how to fool the discriminator, thus enforcing a common representation of source and target data. 
%Inspired by Generative Adversarial Networks~\cite{GANS}, it was originally proposed for domain adaptation in~\cite{ganin2016domain}. 
%Variational Recurrent Adversarial Deep Domain Adaptation \citep[VRADA, ][]{VRADA} and 
%For these algorithms, the architecture of the classifier and discriminator consists of a stack of fully connected layers. The main difference between them lies in their backbones.
%VRADA relies on a Variational Recurrent Neural Network \citep[VRNN, ][]{VRNNs}, which integrates elements of a Variational Auto-Encoder \cite{VAE} into a Recurrent Neural Network in order to achieve the variability observed in highly structured time series data. 
%CoDATS proposes the use of a 1-dimensional fully convolutional neural network (1D CNN) to build the backbone.
%This architecture significantly reduces training time and offers better results compared to VRADA. 

%Additionally, we propose InceptionDANN {replaces CoDATS' 1D CNN backbone with the Inception one} and InceptionCDAN is based on Conditional DANN~\cite{long2018conditional}, where multilinear conditioning is used to capture the cross-covariance between feature representations and classifier predictions. 

\subsubsection{Contrastive learning}

Contrastive learning aims at leveraging data augmentation in order to learn invariant representations of the underlying data.
\cite{CoTMix} propose Contrastive Domain Adaptation for Time-Series Via Temporal Mixup (CoTMix), where 
%augment time series through a \textit{cross-domain temporal mixup strategy}. This strategy consists in 
new time series are built based on
%two time series from different domains, by computing a convex combination of
the time series of one domain and a moving average over a window of length $2L + 1$ of a sample from the second domain.
%The resulting time series is called \textit{dominant} for the first domain.
%For all $\rmX_i^s \in D_s$ and an associated selected $\rmX_j^t \in D_t$, the source-dominant time series are build as follows for each time step $l$,
For all $i \in [1, n_s]$ and an associated $\rmX^t \in D_t$ for each $i$, the source-dominant time series are build as follows for each time step $l$,
\begin{equation}
    (\rmX_{i+n_s}^{s})_{l} = \alpha (\rmX_i^s)_l + (1-\alpha) \frac{1}{2L}\sum_{m=l-L}^{l+L} (\rmX^t)_{m} \text{ and } \rvy_{i+n_s}^{s} = \rvy_{i}^{s},
\end{equation}
where $\alpha \in ]0.5, 1[$.
%$0.5<\alpha<1$.
Thus, the concatenation of the usual source dataset and the new source-dominant dataset has now  $2n_s$ samples: $\{(\rmX_i^{s}, \rvy_i^{s})\}^{2n_s}_{i=1}$.
%This leads to the generation of two new domains, the source-dominant domain and the target dominant domain.  
%A contrastive loss is then computed differently for source and target, due to the absence of labels in target. 
%In the source domain, let the overall source samples be $\{(\rmX_i^{so}, \rvy_i^{so})\}^{2n_s}_{i=1}$, and their corresponding output probabilities be  $\{\rvp_i^{so}\}^{2n_s}_{i=1}$, with $\rvp_i^{so} = C(E(\rmX_i^{so}))$, assuming that the class label is the same for any two corresponding samples from both domains, i.e. $\rvy_i^{s}= \rvy_i^{sd}$ for any $i\in[n_s]$.
The proposed class-aware contrastive loss in source
%is called "class-aware contrastive loss" and
aligns the classifier prediction of the initial and augmented time series belonging to the same class,
\begin{equation}    
    \mathcal{L}_\mathrm{CAC}(E,C) =
    -\sum_{i=1}^{2n_s}\frac{1}{n_{\rvy_i^{s}}-1}\sum_{\substack{j\ne i,\\  y_j^{s}=y_i^{s}}}\log\frac{\exp{(C(E(\rmX_i^{s}))\cdot C(E(\rmX_j^{s}))/\tau)}}{\sum_{a\neq i}\exp{\left(C(E(\rmX_i^{s}))\cdot C(E(\rmX_a^{s}))/\tau\right)}}\;,
\end{equation}
where the symbol $\cdot$ denotes the dot product, $\tau$ is a temperature parameter, and $n_{\rvy_i^{s}}$ is the number of time series with the same class as $\rmX^s_i$.

Similarly, an extended dataset is created for the target domain but with class labels, $\{\rmX_i^{t}\}^{2n_t}_{i=1}$.
%, contrastive learning can only be done in an unsupervised manner. Similarly as in source, the overall target samples are $\{\rmX_i^{to}\}^{2n_t}_{i=1}$, and their corresponding output probabilities are  $\{\rvp_i^{to}\}^{2n_t}_{i=1}$. %, forming a set such that $\rvp_i^{to}= \rvp_i^{t}$ if $i\leq n_t$ and $\rvp_i^{to}= \rvp_i^{td}$ otherwise. 
Thus, the unsupervised contrastive loss aligns the predictions of a sample with its augmented version, only assuming that each new sample belongs to the same class,
\begin{equation}
    \mathcal{L}_\mathrm{UC}(E,C) =- \frac{1}{2 n_t}\sum_{j=1}^{2n_t}\log\frac{\exp{(C(E(\rmX_j^{t}))\cdot C(E(\rmX_{j\pm n_t}^{t}))/\tau)}}{\sum_{a\neq j}\exp{(C(E(\rmX_j^{t}))\cdot C(E(\rmX_a^{t}))/\tau)}},
\end{equation}
where the sign in the notation $j\pm n_t$ depends on whether $j\le n_t$ or $j>n_t$.
%The overall contrastive loss is the sum of the class-aware and the unsupervised contrastive losses: 
%\begin{equation*}
%\mathcal{L}_\mathrm{contrastive} = \mathcal{L}_\mathrm{CAC}(E,C) + \mathcal{L}_\mathrm{UC}(E,C). 
%\end{equation*}
In addition to the usual cross-entropy loss over source domain samples, $\mathcal{L}_C$, CoTMix proposes an entropy loss over target domain samples,
\begin{equation}
    H_t(E,C) = -\frac{1}{n_t}\sum_{j=1}^{n_t} C(E(\rmX_{j}^{t})) \cdot \log C(E(\rmX_{j}^{t})) \;.
    \label{eq:entropy}
\end{equation}
The total loss for the contrastive learning framework is,
\begin{equation}
     \mathcal{L}_\mathrm{total}(E,C) = \mathcal{L}_C(E,C) + H_t(E,C)
     + \lambda \left[\mathcal{L}_\mathrm{CAC}(E,C) + \mathcal{L}_\mathrm{UC}(E,C)\right],
    \label{eq:total_cont_loss}
\end{equation}
where $\lambda$ is a trade-off between the classification and the contrastive loss functions.
%The loss function (\ref{eq:total_cont_loss}) is the learning objective used by the CoTMix and InceptionMix models in our framework.
%In summary, contrastive learning aims at aligning the predictions made by the model for pairs of samples coming from two different domains, thus enforcing domain adaptation.
%CoTMix~\citep{CoTMix} implements contrastive learning between the soft probabilities of source domain samples and their source dominant counterparts and between the soft probabilities of target domain samples and their target dominant counterparts. 
%Source dominant and target dominant samples are produced through the \emph{temporal mixup} operation, which selects a source (or target) domain sample and a target (respectively source) domain sample. 
%The temporal mixup is produced by a component-wise weighted sum between the first sample and a moving average of the second sample. 
%Each temporal mixup should preserve the characteristics of the \emph{dominant} domain while considering the temporal information from the other, \emph{less dominant}, domain.
CoTMix's backbone is a 1D CNN and we proposed InceptionMix that applies the same learning algorithm as CoTMix while employing the Inception backbone.

\subsubsection{Frequency domain analysis} 

% The main proposition of Raincoat~\cite{raincoat} is to use both time and frequency domain features.
\cite{raincoat} propose the Raincoat algorithm, where the main idea is to use both time and frequency domain features.
%This allows the model to learn shifts in the distribution of frequency features that might be vital for the main classification task.
To this purpose, the  model uses a frequency encoder, $E_F:\mathcal{X}\rightarrow\mathcal{Z}_F$, which essentially transforms the input into frequency features with a Fourier Transform followed by learnable convolution~\citep{li2020fourier}.
%, with $\mathcal{Z}_F$ the frequency feature space.
On the other hand, time features are extracted by a traditional 1D CNN backbone $E_T:\mathcal{X}\rightarrow\mathcal{Z}_T$ and the overall encoder is simply the concatenation $E(\rmX) = (E_T(\rmX),E_F(\rmX))$, for any $\rmX\in\mathcal{X}$.
Beyond this key idea, we noticed that the authors’ implementation of RainCoat losses has been updated from what was described in their original paper and continues to evolve over time.
We used the latest version of the implementation available during our study, which is neither the initial version nor the final one, and, importantly, we were able to reproduce the results in the paper.
Thus, what we describe bellow is the exact implementation that we have used.
%we would like to highlight that there is some difference in the losses used between the official repository and the article~\citep{raincoat}.
%We have kept the 

%The Raincoat algorithm aims to align the latent space representations of source and target domain samples. Let $\rmZ^S = \{E(\rmX): \rmX\in D_s\}$ be the set of latent space representations of samples in the source domain, and similarly for the target domain let $\rmZ^T = \{E(\rmX): \rmX\in D_t\}$. The Sinkhorn divergence is computed between the sets $\rmZ^S$ and $\rmZ^T$ and minimized with respect to $E$. Thus, $\mathcal{L}_\mathrm{Sinkhorn}(E) = \mathrm{Sinkhorn}(\rmZ^S,\rmZ^T)$.
To align both domains, Raincoat applies the Sinkhorn divergence~\citep{cuturi2016smoothed} between the source and target distributions pushed forward by the encoder $\mathcal{L}_\mathrm{Sinkhorn}(E) = \mathrm{Sinkhorn}(E_{\#}p_s,E_{\#}p_t)$.
%\begin{equation*}
%    \mathcal{L}_\mathrm{Sinkhorn}(E) = \frac{1}{n_s \times n_t} min_{T \in \mathcal{R}} \sum_{i=1}^{n_s} \sum_{j=1}^{n_t}  \lVert \rmX^s_i - \rmX^s_j \rVert_2^2 + .
%\end{equation*}
%Additionally, Raincoat
%promotes learning of semantic features by minimizing a reconstruction loss. Let 
%proposes a decoder function $G:\mathcal{Z}\rightarrow\mathcal{X}$ which aims at reconstructing the input from its latent space representation, %The reconstruction loss is computed between input samples and their reconstructed counterparts, and it is optimized with respect to the encoder $E$ and the decoder $G$.
%Thus, $\mathcal{L}_R(E,G) = \frac{1}{n_s}\sum_{i=1}^{n_s}d(\rmX_i^s,G(E(\rmX_i^s))) + \frac{1}{n_t}\sum_{i=1}^{n_t}d(\rmX_i^t,G(E(\rmX_i^t)))$, with $d$ some distance in $\mathcal{X}$.
%\tanguy{if someone know already what is this contrative loss used}
Additionally, to enforce a latent space where same class source points are close to each other and separate from the other classes, a contrastive loss is used,
% \mathbb{1}_{\rvy^s_i == \rvy^s_j}
\begin{equation}
    \mathcal{L}_R(E) = \frac{1}{2} \sum_{i=1}^{n_s} \sum_{\substack{j=1\\ \rvy^s_i = \rvy^s_j}}^{n_s}  \lVert E\left(\rmX^s_i\right) - E\left(\rmX^s_j\right) \rVert_2^2 + \sum_{\substack{j=1\\ \rvy^s_i \neq \rvy^s_j}}^{n_s}  \max\left[0, \frac{1}{2} - \lVert E\left(\rmX^s_i\right) - E\left(\rmX^s_j\right) \rVert_2\right]^2.
\end{equation}
Thus, the total loss function used in the Raincoat algorithm is,
\begin{equation}
    \mathcal{L}_\mathrm{total}(E,C) = \mathcal{L}_C(E, C) + \mathcal{L}_\mathrm{Sinkhorn}(E) + \mathcal{L}_R(E)\,.
\end{equation}

%In summary, Raincoat is a novel method for UDA in time series proposed by~\cite{raincoat} in which the temporal features and frequency features of the time series are analyzed separately. Raincoat consists of three different modules: an encoder, a decoder and a classifier. The main novelty is in the encoder, which treats separately time and frequency features. Time features are analyzed by a traditional backbone, such as a 1D CNN, while the frequency features are extracted by: (1) smoothing, (2) applying a discrete Fourier transform, (3) multiplying by a learned weight matrix, (4) transforming into amplitude and phase components, and (5) concatenating. Once the frequency and time features are extracted, these are concatenated and passed on to a classifier which is trained for the main classification task.
%The role of the decoder is to reconstruct the input samples from the features extracted by the encoder. A reconstruction loss computed over both source and target domain samples forces the encoder to produce an accurate representation of the target samples. Additionally, the Sinkhorn divergence between the extracted source domain and target domain features is computed and used to align the target and source domain representations. 
InceptionRain also proposed in this work, uses the Inception backbone as a time feature encoder for the Raincoat algorithm.

%\subsubsection{Non-deep UDA}
%In addition to the algorithms mentioned above, we consider Optimal Transport Domain Adaptation \citep[OTDA, ][]{OTDA} as a non-deep approach, which consists in aligning the source distribution over the target one using the Optimal Transport plan.
%Note however that this approach is not originally designed for time series. \todo{[keep it only if we add subspace alignment and KMM, otherwise we will drop it]}

\subsection{Tuning models without labels}
\label{sec:model-selection}
The aim of hyperparameter tuning for DA is to find the hyperparameters that lead to a model $f$ minimizing the target risk $\mathcal{R}_t(f)$.
This is particularly challenging in unsupervised DA, given the absence of labels in the target domain for model evaluation, and can drastically impact the methods’ performances~\citep{Musgrave2022ThreeNew}.
While it is a common practice in many UDA papers to manually set hyperparameters for each scenario~\citep{tzeng2017adversarial,CoDATS}, three standard approaches for an automatic selection of hyperparameters are considered in the benchmark.
%\tanguy{Still issue with notation, T or t ? s or S ?}
\subsubsection{Target Risk} Several methods~\citep{long2015learning,OTDA,saito2018maximum} rely on the empirical target risk\footnote{\label{footnote1}Referred to as target/source risk for short in the sequel.} to select models, computed using target labels,
\begin{equation}
\hat{\mathcal{R}}_t(f) = \frac{1}{n_t} \sum_{j=1}^{n_t} \Ls (\rvy_j^t,f(\rmX_j^t)).
\end{equation}
If $\Ls$ is the $0-1$ loss, then $\hat{\mathcal{R}}_t$ is the target accuracy.
However, this method is questionable because it will not be applicable for real UDA experiments and serves more as an oracle and an upper bound on performance.

\subsubsection{Source Risk}
In \cite{ganin2015unsupervised} authors select hyperparameters by using the empirical source risk\hyperref[footnote1]{$^1$}:
\begin{equation}
\hat{\mathcal{R}}_s(f) = \frac{1}{n_s} \sum_{i=1}^{n_s} \Ls (\rvy_i^s,f(\rmX_i^s)).
\end{equation}
Although it should lead in theory to a misestimation of the target risk in the presence of a large domain gap, 
\cite{Musgrave2022ThreeNew} have shown that it results in good performance on UDA for computer vision.

\subsubsection{Importance Weighted Cross Validation (IWCV)}
IWCV is a theoretically sound approach to estimate the target risk, proposed by \cite{sugiyama2007covariate} and applied in \cite{CoTMix} and \cite{long2018conditional}. 
Each source sample $\rmX^s_i$ is weighted by the ratio between the probability density of the target %$p_t(\rmX)$ and the source $p_s(\rmX)$.
$p_t(X)$ and the source $p_s(X)$ at point $\rmX^s_i$.
The authors prove that the following equality holds under
the covariate shift assumption,
for any classifier $f$,
%\[ \mathcal{R}_t (f)= \mathbb{E}_{(\rmX,\rvy)\sim p_s}\left[ \frac{p_t(\rmX)}{p_s(\rmX)} \Ls (\rvy,f(\rmX))\right]. \]
\begin{equation}
\mathcal{R}_t (f)= \mathbb{E}_{(X,y)\sim p_s}\left[ \frac{p_t(X)}{p_s(X)} \Ls (y,f(X))\right].
\end{equation}
Taking the following empirical expectation leads to a proxy of the target risk: %.
\begin{equation}
IWCV(f) = \frac{1}{n_s} \sum_{i=1}^{n_s} \frac{\hat{p}_t(\rmX_i^s)}{\hat{p}_s(\rmX_i^s)}\Ls (\rvy_i^s,f(\rmX_i^s)).
\end{equation}
In practice, the performance of IWCV is strongly limited by the unverifiability of the covariate shift assumption and the difficulty of estimating the densities, especially for time series data.

Although other methods exist in the literature %, such as 
\citep{zhong2010cross,you2019towards,robbiano2022adversarial,chuang2020estimating,saito2021tune,Musgrave2022ThreeNew}, we choose the approaches that offer the best practicality, robustness, lowest computational cost, and are well studied in the literature.

\subsection{Estimating shift from source to target}

Gaining insights into the performance of the UDA algorithms based on the shift level between source and target provides valuable understanding of their behavior on a dataset.
Yet, estimating the shift between two datasets is a challenging area of research, particularly in multivariate time series with the presence of shifts in both the temporal and feature spaces. 

%To the best of our knowledge, no existing distances or metric between distributions take both feature and temporal shift simultaneously: Dynamic Time Warping (DTW) \cite{berndt1994using} and its multivariate variants \citep[and references therein]{shifaz2023elastic} only consider the temporal and feature shift separately, Wasserstein distance~\cite{redko2017theoretical} on the contrary mainly focuses on the feature shift and do not consider the temporal shift, and Maximum Mean Discrepancy (MMD)~\cite{long2017deep,sinn2012detecting} has been applied to either types of shift, but not both simultaneously.
%However, our experiments clearly show that the best method for time series adaptation is RainCoat, which considers both types of shift through the preprocessing and architecture. 
%Further work is thus needed to estimate both types of shift at the same time. 
%Additionally, the aforementioned approaches can be tedious to estimate for each pair of time series with a complexity of $\mathcal{O}(n^2)$ in most cases.

%To overcome both issues and still provide some insight on how UDA approaches perform to various levels of shift,
We propose to estimate this shift by computing a shift proxy $SP$ as the relative loss in accuracy between source and target data for the Inception classifier,
\begin{equation}\label{eq:shift-proxy}
SP(D_s, D_t) = \frac{\hat{\mathcal{R}}_s(f_{Inception}) -\hat{\mathcal{R}}_t(f_{Inception})}{\hat{\mathcal{R}}_s(f_{Inception})}.
\end{equation}
Inception does not use any adaptation technique, therefore, if there is no shift in the data, it should perform similarly in source and target. On the contrary, when the shift increases, this method fails to get a correct classification in the target domain, as data becomes too different to what it is has been trained on.
We should note that since this shift estimation technique requires prior knowledge of target labels, it is only useful for analytical purposes and cannot be applicable in a real world UDA scenario.
Several other methods exist for an analysis solely based on either temporal or feature shift, notably Dynamic Time Warping~\citep{berndt1994using,shifaz2023elastic}, Wasserstein distance~\citep{redko2017theoretical} and Maximum Mean Discrepancy~\citep{long2017deep,sinn2012detecting}.

\section{Experimental setup}
In this section we provide details about the datasets used in this benchmark followed by a thorough description of the pipeline employed for evaluation.

\subsection{Datasets used for evaluation}
\subsubsection{Overview of the used datasets}
Table~\ref{tab:datasets_stats} presents an overview of each dataset, including  statistics such as number of domains, number of classes and channels, length of each time series, and its theme out of the four following ones: machinery, motion, medical and remote sensing.

\begin{table*}[ht!]
    \centering
    \caption{Description of unsupervised domain adaptation time series datasets. Bold indicates the \textbf{datasets new to UDA} proposed in this benchmark.}
    \label{tab:datasets_stats}
    \small
    \begin{tabular}{ l  l l  l  l  l }
    \toprule
        Dataset & Domains & Classes & Length & Channels & Theme \\ 
        \midrule
		\textbf{bearing} & 4 & 4 & 512 & 1 & machinery \\
		\textbf{ecg}& 3 & 5 & 1000 & 12 & medical  \\
		\textbf{equations} & 2 & 15 & 64 & 13 & motion\\
        \textbf{ford} & 2 & 2 & 500 & 1 & machinery \\
        har & 30 & 6 &  128 & 9 & motion \\  
		hhar& 9 & 6 & 128  & 3 & motion \\ 
        mfd  & 4 & 3 & 5120 & 1 & machinery \\ 
        \textbf{miniTime} & 4 & 8 & 39 & 10 & remote sensing\\  
        \textbf{muscle} & 8 & 2 & 3000 & 1 & medical \\
        sleep & 20 & 5 & 3000 & 1 & medical\\
        \textbf{sports} & 8 & 19 & 125 & 45 & motion \\ 
        wisdm & 36 & 6 & 128 & 6 & motion \\ 
        \bottomrule
    \end{tabular}
\end{table*}

The most commonly used datasets in UDA for TSC \citep{anguita2013public,stisen2015smart,lessmeier2016condition,ragab2023adatime,kwapisz2011activity} revolve around human activity recognition, fault detection and sleep stage prediction: (1) Human Activity Recognition (HAR); (2) Heterogeneity Human Activity Recognition (HHAR); (3) Machine Fault Diagnosis (MFD); (4) Sleep Stage; (5) Wireless Sensor Data Mining (WISDM).
In an effort to increase the diversity in terms of applications as well as having a larger sample for comparing the classifiers, we introduce seven datasets from TSC and tailor them into a DA setting,
highlighted in bold in Table~\ref{tab:datasets_stats}: Ford, CWR Bearing (bearing), PTB XL ECG (ecg), Ultrasound Muscle Contraction (muscle), Online Handwritten Equations (equations), Sport Activities (sports), and Mini Time Match (miniTime). 
The added datasets are accompanied by the originally specified or commonly used preprocessing. 
We have ensured the presence of a shift between source and target domain by running Inception and observing the drop in accuracy when training on source and evaluating on target. 
For datasets where the number of possible UDA scenarios exceeds five, we have taken a random selection of five scenarios to limit the number of experiments which increases exponentially with each additional scenario.

\begin{figure}[!ht]
    \centering
    \begin{subfigure}{0.49\linewidth}
    \centering
    \includegraphics[width=.95\linewidth]{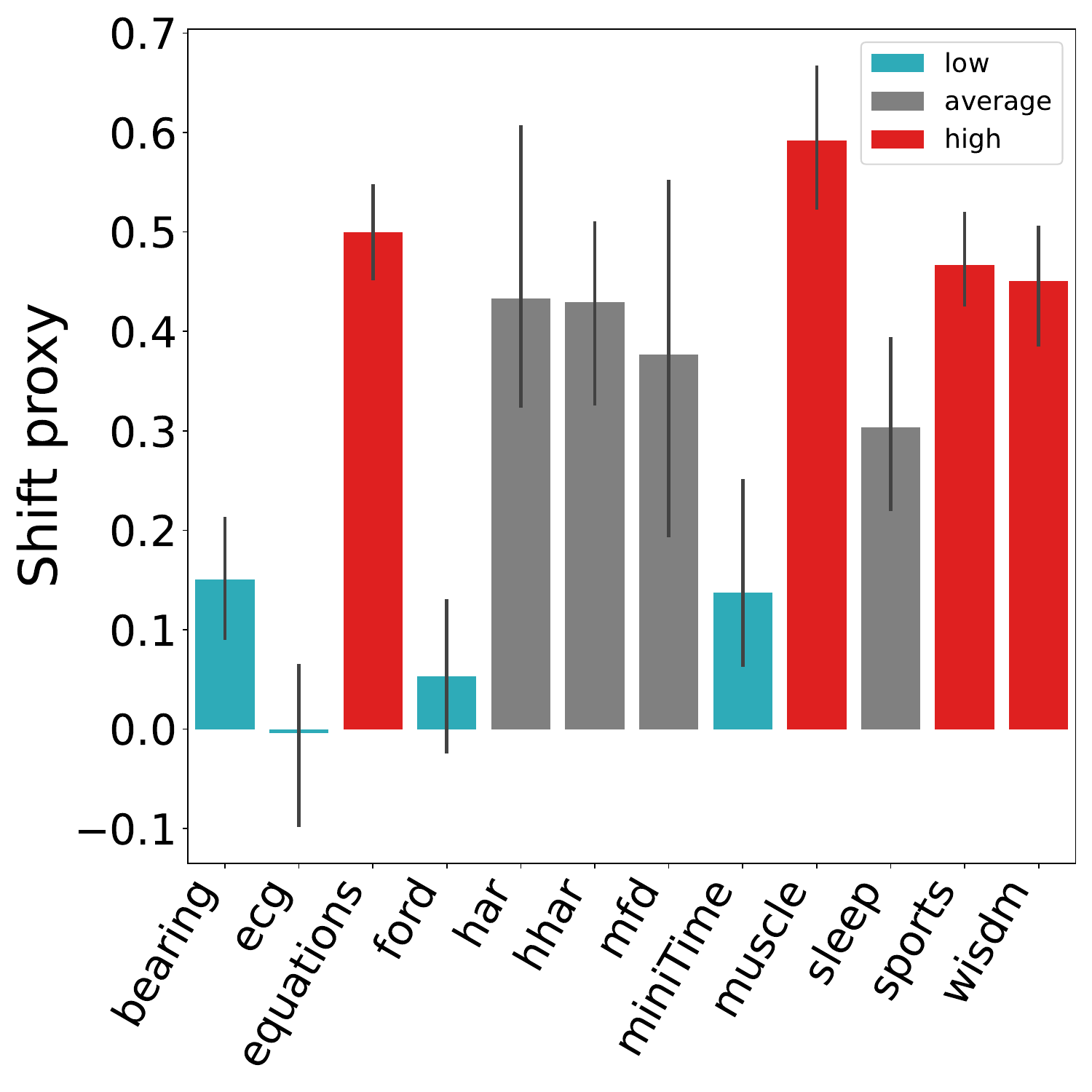}
    \caption{Shift estimation $SP$}
    \end{subfigure}
    \begin{subfigure}{0.49\linewidth}
        \centering
    \includegraphics[width=.95\linewidth]{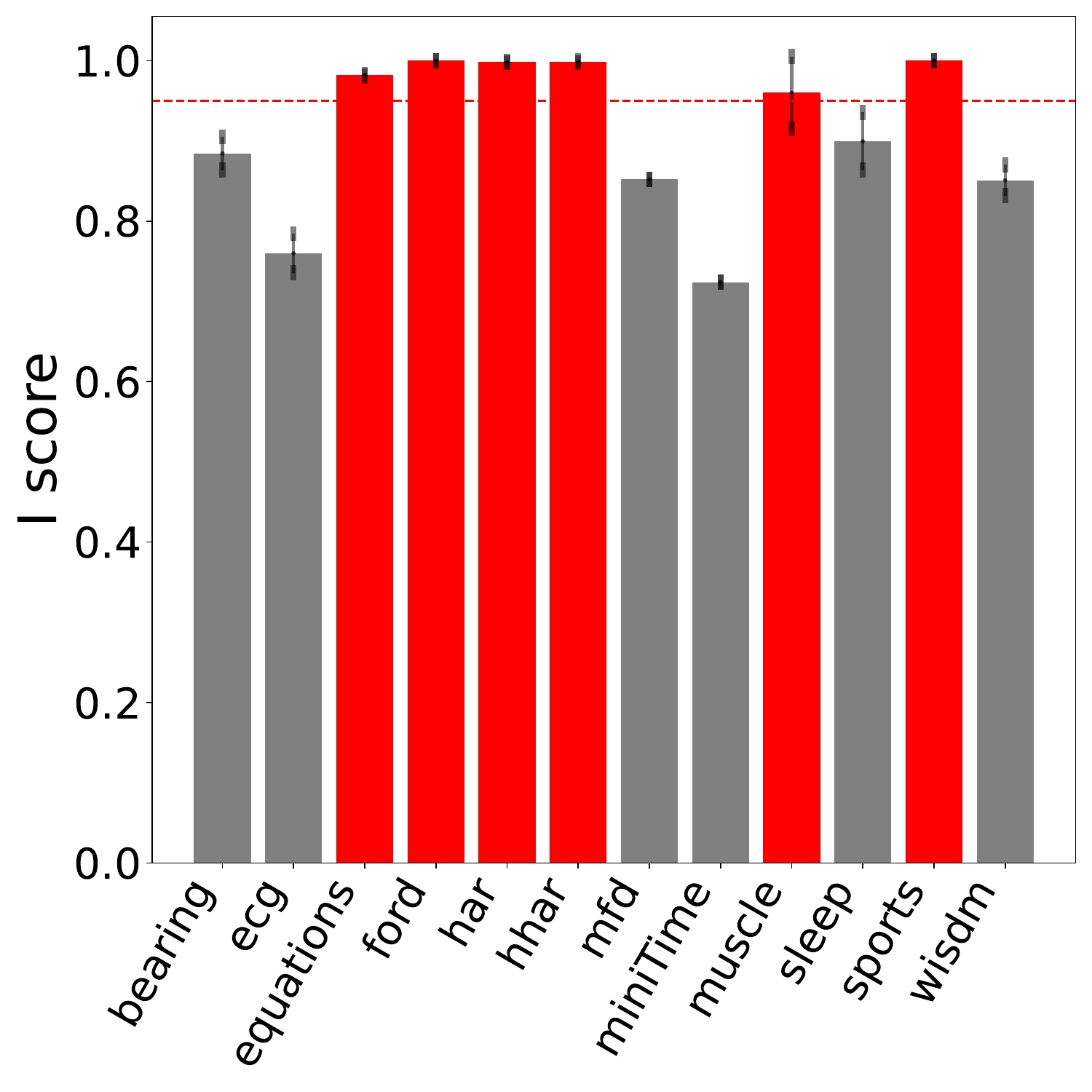}
    \caption{Imbalance score $I$}
    \label{fig:i_score}
    \end{subfigure}
    \caption{Repartition of the shift estimation and imbalance score of the different datasets and pairs.}
    \label{fig:datasets-characteristics}
\end{figure}

As demonstrated by Table~\ref{tab:datasets_stats}, the datasets proposed in this study are thus diverse in terms of number of classes and channels, length of the time series and themes.
Fig.~\ref{fig:datasets-characteristics} additionally shows their  diversity in terms of shift between pairs of datasets, estimated through the shift proxy in  (\ref{eq:shift-proxy}), and the imbalance score based on the Shannon entropy scaled between $0$ and $1$ by dividing by the log number of classes $K$,
%, a measure of uncertainty or information content in a dataset:
\begin{equation}
    I = -\frac{1}{log(K)}\sum_{k=1}^{K} p_k log(p_k),
\end{equation} where, $p_k$ is the proportion of instances belonging to the class $k$ in a dataset.
%The computation of $p_k$ for each class $k$  is given by $ p_k = \frac{n_{k}}{n}$, where $n$ is the total number of samples and $n_{k}$ is the number of samples in the class $k$.
%To gain better insights, we propose normalizing the entropy to a range between 0 and 1 by dividing it by $log(K)$.
%Thus, $ I = \frac{H}{log(K)}.$
We define a threshold, categorizing datasets as highly imbalanced if their $I$ scores is lower than 0.95. 

%explained in Appendix~\ref{app:dataset_imbalance}. 

%For completeness, the following is a detailed description of the datasets used in this benchmark.
\subsubsection{Detailed description of the used datasets}
\paragraph{Human Activity Recognition (har)}
HAR is the most common time series dataset~\citep{anguita2013public} available on UCI repository. 
It was collected from 30 subjects to classify their activities based on 3 sensors namely, accelerometer, gyroscope, and body sensors. 
Each subject has performed six activities, i.e., walking, walking upstairs, downstairs, standing, sitting, and lying down. Given the diversity observed among subjects, we consider each individual subject as an independent domain. 

\paragraph{Heterogeneity Human Activity Recognition (hhar)}
HHAR is also a very common time series problem~\citep{stisen2015smart} available on UCI. It was collected from 9 subjects to capture their activities: Biking, Sitting, Standing, Walking, Stair Up and Stair down. 
The activity is recorded by sensors coming from 2 devices: smartphones (2 Samsung Galaxy S3 mini, 2 Samsung Galaxy S3, 2 LG Nexus 4, 2 Samsung Galaxy S+) and smart watches (2 LG watches, 2 Samsung Galaxy Gears). 
Similarly to the HAR dataset, we consider each individual as an independent domain for the underlying UDA task. 

\paragraph{Machine Fault Diagnosis (mfd)}
MFD~\citep{lessmeier2016condition} was proposed by Paderborn University to detect early-stage machine faults through the analysis of vibration signals. 
The TSC task consists of predicting three possible classes: normal, inner and outer damages.
The dataset was gathered across four distinct operational conditions, each one considered as an independent domain. 

\paragraph{Sleep Stage (sleep)}
Sleep Stage~\citep{ragab2023adatime} consists of classifying the electroencephalography (EEG) signals into five labels: Wake, Non-Rapid Eye Movement stages and Rapid Eye Movement.
Scenarios are split based on the various patients present in the dataset.

\paragraph{Wireless Sensor Data Mining (wisdm)}
WISDM is an other collection of sensor data commonly used for time series research in the field of Human Activity Recognition~\citep{kwapisz2011activity}. 
It includes data from accelerometers and gyroscopes embedded in smartphones to capture different human activities: walking, jogging, going up and down stairs, sitting, and standing. 
It was collected on 36 different subjects, each one considered as an independent domain.

\paragraph{Ford (ford)}
Ford~\citep{dau2019ucr} was initially employed in the IEEE World Congress on Computational Intelligence in 2008 as part of a competition. 
The primary classification task involved diagnosing specific symptoms within an automotive subsystem. 
Each instance in the dataset comprises 500 measurements of engine noise alongside a corresponding classification label of whether or not an anomaly exists in the engine. 
There are two distinct domains: (1) FordA where data were gathered under typical operational conditions, characterized by minimal noise interference; (2) FordB where data were collected under noisy conditions.

\paragraph{CWR Bearing (bearing)}
CWR Bearing~\citep{zhang2019machine} consists of motor vibration data collected using accelerometers placed at the 12 o’clock position at both the drive end and fan end of the motor housing in order to detect normal and faulty bearings, with one normal class and 3 fault classes. 
The data was collected at 12,000 %samples per second and at
and 48,000 samples per second for drive end bearing experiments.  
Five distinct scenarios are generated from different motor conditions.

\paragraph{PTB XL ECG (ecg)}
PTB~\citep{ptbXLecg} is a collection of 549 high-resolution 15 channels ECGs. 
The data was recorded at different clinical sites and gathers
294 subjects, including healthy subjects as well as patients with a variety of heart diseases representing the 5 different classes.
Five domain adaptation scenarios are created from 4 clinical sites. 

\paragraph{Ultrasound Muscle Contraction (muscle)}
UMC~\citep{brausch2022classifying} is a collection of 21 sets of one-dimensional ultrasound raw radio frequency data (A-Scans) measured on calf muscles of 8 healthy volunteers, each A-Scan consisting of 3000 amplitude values. 
The purpose is to determine whether the muscle is contracted or not.
Each individual subject is considered as an independent domain.

\paragraph{Online Handwritten Equations (equations)} 
OnHWeq~\citep{OnHWeq} consists of recognizing handwritten equations based on multivariate time series data captured from sensors placed on a sensor enhanced pen.
With 12 possible equations and 55 different writers, the time series classification task is based on 12 labels while domain adaptation scenarios are split based on the writer's identity.
 
\paragraph{Sport Activities (sport)}
Sport Activities~\citep{altun2010comparative} comprises motion sensor data of 19 daily sports activities performed by 8 subjects in their own style for 5 minutes. 
Data was recorded using five Xsens MTx sensor units placed on the torso, arms, and legs. 
Each individual subject is considered as an independent domain.
 
\paragraph{MiniTimeMatch (miniTime)}
The miniTimeMatch 
~\citep{nyborg2022timematch} dataset is a crop-type mapping that covers four areas across Europe, located in Austria, Denmark, South and North of France. 
It comprises a series of time-stamped multi-spectral measurements derived from satellite imagery captured at specific geographic coordinates.
The aim is to recognize the type of crop of a parcel among 8 categories, such as corn or wheat. 
We considered the four regions as independent domain
each region being an independent domain.

\subsection{Benchmarking pipeline} 
\label{section:xps_setup}

Comparing UDA algorithms for TSC is challenging due to the number of scenarios and variations in preprocessing, time splitting, and hyperparameter selection among different UDA approaches in the literature.
To address these challenges, we propose a unified framework that facilitates fair and equitable comparisons of UDA algorithms for TSC.
Its scalability and high modularity enables the incorporation of new datasets, preprocessing methods, UDA algorithms and hyperparameter tuning approaches.

Our benchmarking framework consists of five stages, similarly to traditional machine learning pipeline.
(1) \textbf{Loading}: raw time series datasets are downloaded from their original source %that we 
separated into train, validation and test sets if not already split.
Labels from the source domain are utilized to %ensure an equitable 
preserve the distribution of classes across all three sets.
In line with supervised ML conventions, the class proportions between the test set and the training/validation sets remain consistent in the target domain. However the stratification is not performed when splitting the train and validation target set, where labels are not supposed to be available.
Additionally, when possible, temporal causality is ensured for all splits.
(2) \textbf{Preprocessing}: each raw time series is preprocessed in the same way for all methods and following the recommendations of the paper that proposed the dataset (e.g. z-normalization, resampling, interpolation.%, see details in Appendix~\ref{app:hyperparameters}).
(3) \textbf{Tuning}: for each couple of dataset-classifier, we search for the best hyperparameters using the three methods presented in Section \ref{sec:model-selection}. 
The marginal distributions %$p_t(\rmX)$ and $p_s(\rmX)$ 
$p_t(X)$ and $p_s(X)$ in IWCV are estimated by a mixture of 5 Gaussian components and $\Ls$ is taken as the cross-entropy.
For all 3 methods, the Source or Target Risk is estimated using the validation set, which is unseen during training.
(4) \textbf{Training}: for each of the three model selection methods, we take the best hyperparameter set and re-train the model, check-pointing the weights at each epoch using the same metric and validation set used during tuning. 
(5) \textbf{Evaluation}: for each trained model, we evaluate the final metric over the held-out test set (source and target). 
Thus, contrary to some DA settings, the test data are never seen during training nor validation (except when tuning with Target Risk).

To ensure fairness across all algorithms, we fixed the hyperparameter tuning budget for all experiments to 12 hours of GPU time, and the training budget to 2 hours of GPU time for each model on a given dataset for a given set of hyperparameters. 
Thus, the total sequential runtime of this benchmark is approximately $8748$ hours, corresponding to almost one year.
The benchmark framework code is publicly available on GitHub\footnote{\url{https://github.com/EricssonResearch/UDA-4-TSC}} in order to ensure the reproducibility of our work.

%is detailed in Appendix~\ref{app:framework} and the code is publicly available on GitHub\footnote{\url{https://github.com/EricssonResearch/UDA-4-TSC}} in order to ensure the reproducibility of our work. 

\section{Results and analysis of the benchmark}

In the following subsections we start by presenting an overall comparison of classifiers over multiple datasets. 
Then, we perform a statistical analysis between the different neural network backbones and the different hyperparameter tuning methods. 
We also perform a meta-analysis for understanding how data characteristics impact the performance of UDA algorithms.
% Deeper insights are provided in Appendices~\ref{app:furth_analysis} and \ref{app:meta-analysis}, and the complete results for each dataset in Appendix~\ref{app:acc_dataset}.
Note that some of our results differ in accuracy  from the original papers, which presented the following limitations:
(1) hard-coded hyperparameters were set for a given scenario~\citep{raincoat}; or (2) target labels were used
for tuning the hyperparameters~\citep{CoDATS}; or (3) the time series segmentation did not respect temporal causality thus introducing data leakage~\citep{CoTMix}.

\subsection{Comparison of classifiers over all datasets}\label{subsec:clf_dts}

Following the classical approach for comparing multiple time series classifiers~\citep{bagnall2017great}, 
we first perform a Friedman test~\citep{friedman1940comparison} over all the algorithms which rejects the null hypothesis over the different unsupervised domain adaptation scenarios.
The benchmark average ranking are visualized via critical difference diagrams~\citep{demsar2006statistical}, with the  
omission of the post-hoc analysis based on Wilcoxon-signed rank test with Holm's alpha correction due to its artifacts~\citep{lines2018time}.

\begin{figure}[ht]
    \centering
    \includegraphics[width=\linewidth]{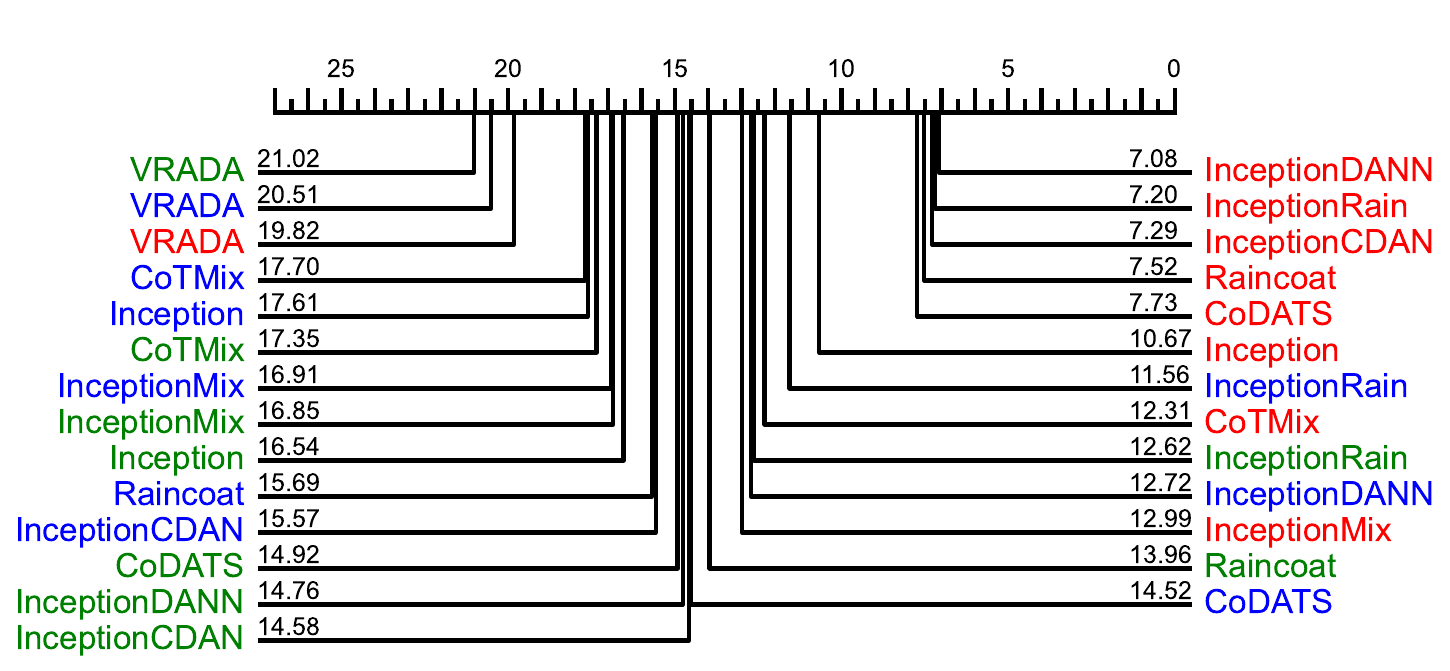}
    \caption{Average rank diagrams based on the accuracy of all UDA algorithms and hyperparameter tuning methods. The colors specify the hyperparameter tuning method: green for \textcolor{vertFonce}{IWCV}, blue for \textcolor{blue}{Source Risk} and red for \textcolor{red}{Target Risk}.}
    \label{fig:all_algo_hyp}
\end{figure}

Fig.~\ref{fig:all_algo_hyp} showcases the average rank of all classifiers and all hyperparameter tuning approaches, where colors distinguish hyperparameter tuning methods: IWCV is represented in green, Source Risk in blue, and Target Risk in red. Table~\ref{tab:acc_summary_clf} displays the average accuracies per hyperparameter tuning approach for each UDA algorithm.

%Violinplots of the accuracy and F1-score of each UDA algorithm and hyperparameter tuning are provided in Appendix~\ref{app:furth_analysis_dist_acc} (Fig. \ref{fig:hyperparmCom_acc}).
%From all these visualizations, we can extract the following statements.

\begin{table}[ht]
\small
\centering
\caption{Average accuracy over all datasets for the three model selection methods: Source Risk, IWCV and Target Risk.}
\label{tab:acc_summary_clf}
\begin{tabular}{l | c c | c}
    \toprule
   Classifier &            IWCV &     Source Risk &     Target Risk \\ \hline %\midrule
  Source only & 0.59 $\pm$ 0.07 &  0.60 $\pm$ 0.06 & 0.70 $\pm$ 0.05 \\ \midrule
       CoDATS & 0.64 $\pm$ 0.07 & 0.66 $\pm$ 0.05 & 0.76 $\pm$ 0.04 \\
       CoTMix & 0.57 $\pm$ 0.05 & 0.57 $\pm$ 0.04 & 0.66 $\pm$ 0.03 \\
InceptionCDAN & 0.64 $\pm$ 0.05 & 0.63 $\pm$ 0.04 & 0.75 $\pm$ 0.06 \\
InceptionDANN & 0.63 $\pm$ 0.06 & 0.67 $\pm$ 0.06 & 0.75 $\pm$ 0.05 \\
 InceptionMix & 0.56 $\pm$ 0.07 &  0.57 $\pm$ 0.10 & 0.63 $\pm$ 0.08 \\
\textbf{InceptionRain} & \textbf{0.67 $\pm$ 0.06} & \textbf{0.68 $\pm$ 0.07} & \textbf{0.77 $\pm$ 0.04} \\
%         OTDA & 0.48 $\pm$ 0.06 &  0.5 $\pm$ 0.04 & 0.53 $\pm$ 0.03 \\
     Raincoat & 0.66 $\pm$ 0.06 & 0.65 $\pm$ 0.06 & 0.76 $\pm$ 0.05 \\
%         SASA & 0.45 $\pm$ 0.03 & 0.48 $\pm$ 0.02 &  0.5 $\pm$ 0.03 \\
        VRADA &  0.50 $\pm$ 0.05 &  0.50 $\pm$ 0.05 & 0.52 $\pm$ 0.06 \\ \hline %\midrule
  
  Target only &  0.90 $\pm$ 0.04 &  0.90 $\pm$ 0.0 &  0.90 $\pm$ 0.04  \\
 \bottomrule
\end{tabular}
\end{table}

% \begin{table}[ht]
% \small
% \centering
% \caption{Accuracy results summary table evaluated the 3 model selection methods: Source Risk, IWCV and Target Risk.}
% \label{tab:acc_summary_clf}
% \begin{tabular}{l  c c c}
%     \toprule
%    Classifier &            IWCV &     Source Risk &     Target Risk \\ \midrule
%   \textit{Source only} & \textit{0.59 $\pm$ 0.07} &  \textit{0.60 $\pm$ 0.06} &  \textit{0.70 $\pm$ 0.05} \\ %\midrule
%   \\[-0.7em]
%        CoDATS & 0.64 $\pm$ 0.07 & 0.66 $\pm$ 0.05 & 0.76 $\pm$ 0.04 \\
%        CoTMix & 0.57 $\pm$ 0.05 & 0.57 $\pm$ 0.04 & 0.66 $\pm$ 0.03 \\
% InceptionCDAN & 0.64 $\pm$ 0.05 & 0.63 $\pm$ 0.04 & 0.75 $\pm$ 0.06 \\
% InceptionDANN & 0.63 $\pm$ 0.06 & 0.67 $\pm$ 0.06 & 0.75 $\pm$ 0.05 \\
%  InceptionMix & 0.56 $\pm$ 0.07 &  0.57 $\pm$ 0.10 & 0.63 $\pm$ 0.08 \\
% \textbf{InceptionRain} & \textbf{0.67 $\pm$ 0.06} & \textbf{0.68 $\pm$ 0.07} & \textbf{0.77 $\pm$ 0.04} \\
% %         OTDA & 0.48 $\pm$ 0.06 &  0.5 $\pm$ 0.04 & 0.53 $\pm$ 0.03 \\
%      Raincoat & 0.66 $\pm$ 0.06 & 0.65 $\pm$ 0.06 & 0.76 $\pm$ 0.05 \\
% %         SASA & 0.45 $\pm$ 0.03 & 0.48 $\pm$ 0.02 &  0.5 $\pm$ 0.03 \\
%         VRADA &  0.50 $\pm$ 0.05 &  0.50 $\pm$ 0.05 & 0.52 $\pm$ 0.06 \\ %\midrule
%   \\[-0.7em]
%   \textit{Target only} &  \textit{0.90 $\pm$ 0.04} &  \textit{0.90 $\pm$ 0.04} &  \textit{0.90 $\pm$ 0.04} \\
%   \bottomrule
% \end{tabular}
% \end{table}

%From results presented in Fig.~\ref{fig:all_algo_hyp} and in Table~\ref{tab:acc_summary_clf}, we highlight three main points:

\paragraph{Best accuracy achieved by CNN-based and frequency analysis backbones}

The best-performing approaches are Raincoat and CoDATS, along with their versions using the Inception backbone (InceptionRain, InceptionDANN, Inception-CDAN). These algorithms' backbones rely on temporal convolutions and/or the extraction of pertinent time series features such as frequency, amplitude, and phase. Thus, the selection of a backbone architecture that effectively incorporates temporality is essential for UDA in TSC. Additionally, Fig.~\ref{fig:all_algo_hyp} shows that InceptionRain has the highest average rank among all 9 models when focusing on Source Risk and IWCV, while InceptionDANN ranks slightly higher with Target Risk, and also obtains the best average accuracy for all hyperparameter tuning approaches, as shown in Table~\ref{tab:acc_summary_clf}.

\paragraph{Some UDA approaches perform worse than no DA}
On the contrary, we  notice on Table~\ref{tab:acc_summary_clf} that VRADA, CoTMix and InceptionMix perform worse than training a good classifier on source only and applying it in target with no adaptation, with a difference in accuracy around 0.1 for VRADA and 0.03 for CoTMix and InceptionMix. This is expected for VRADA as it relies on an RNN architecture which has been shown to underperform in classification tasks compared to its convolutional counterpart \citep{smirnov2018time}.
%The memory component of an RNN architecture makes it more relevant in anomaly detection than in more generic classification.
This suggests that the adversarial learning strategy is not sufficient for good results, the choice of a backbone is important as well. As for CoTMix-based approaches, %our benchmark results suggest that the contrastive loss in CoTMix is not able to generalize well. A 
a possible explanation for their performance could be that the temporal mixup strategy mixes time series from different classes, due to the absence of labels in target. Therefore, the resulting augmented dataset tends to blur the decision frontiers between classes instead of improving their separation. 

\paragraph{Similar performance of IWCV and Source Risk in UDA, but the road toward good hyperparameter tuning in UDA is still long} Table~\ref{tab:acc_summary_clf} displays very similar performance when hyperparameters are tuned with IWCV and Source risk, and the ranking of UDA algorithms tuned with both hyperparameter tuning approaches are spread evenly across Fig.~\ref{fig:all_algo_hyp}. Still, their performance are significantly lower than those with Target risk (if target labels were available), %, as shown in Table~\ref{tab:acc_summary_clf} and on  Fig.~\ref{fig:all_algo_hyp} 
as shown with the gap between top-ranked methods (rightmost cluster on  Fig.~\ref{fig:all_algo_hyp}), including algorithms that have all been tuned with Target Risk, and the other approaches. 
This is expected since target labels are utilized to choose the optimal hyperparameters, which is impractical.
We also notice that the intra-cluster distance, in terms of average rank, is significantly lower when using the Target Risk for hyperparameter tuning. 
The latter observation suggests that a good hyperparameter tuning method can close the gap between several classifiers, which is in line with the recent domain adaptation reviews for computer vision~\citep{musgrave2021unsupervised}.

\subsection{Comparison of backbones} 
\label{section:backbones}

\begin{figure}[!ht]
    \centering     
    \begin{subfigure}{0.49\linewidth}
         \centering
         \includegraphics[width=\linewidth]{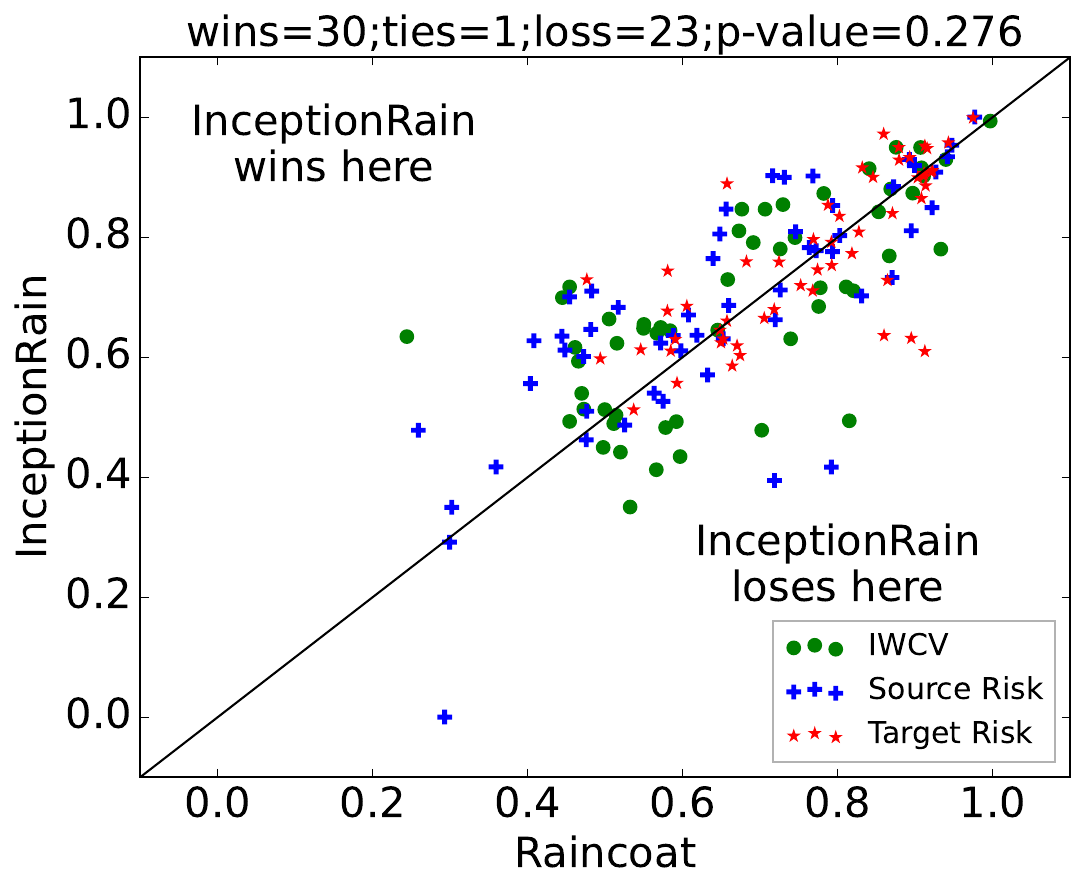}
         \caption{InceptionRain vs Raincoat}
         \label{fig:top1-da-method_paper}
     \end{subfigure}
    \begin{subfigure}{0.49\linewidth}
         \centering
         \includegraphics[width=\linewidth]{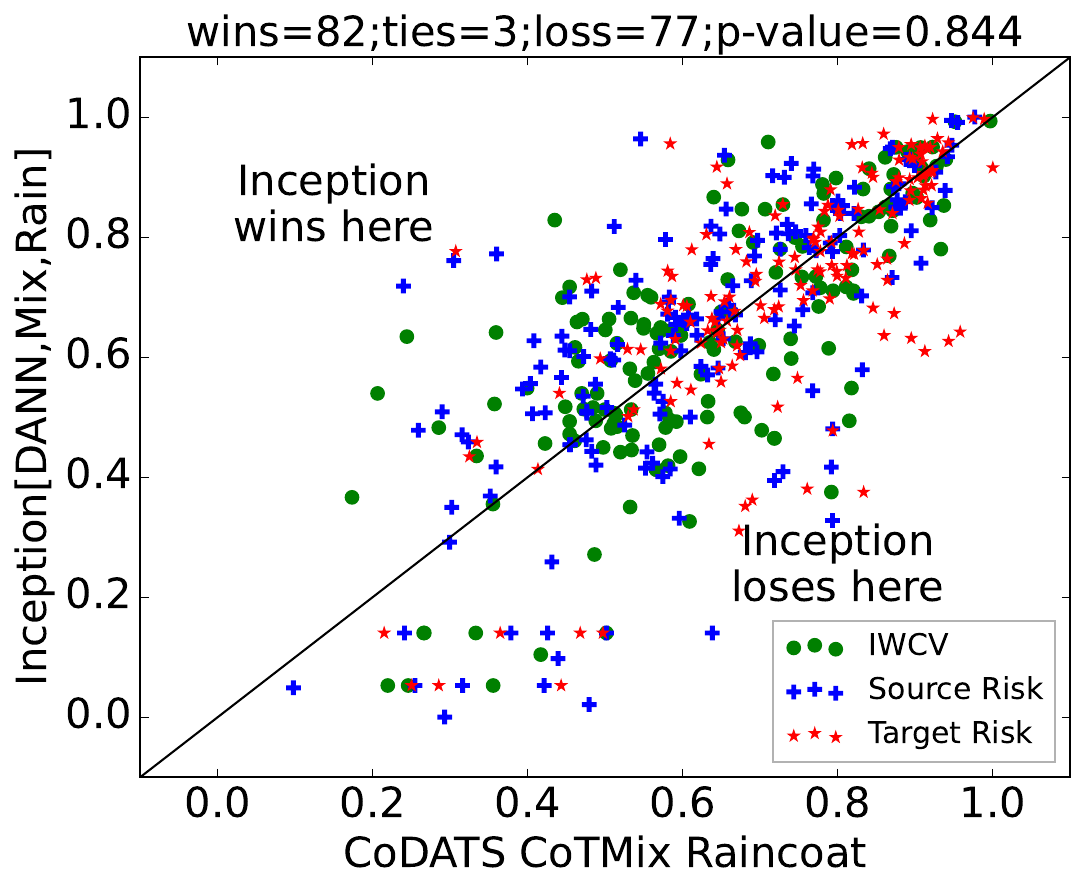}
         \caption{Inception vs original backbones}
         \label{fig:all-da-method}
     \end{subfigure}
\caption{Pairwise accuracy comparison of InceptionRain against Raincoat and of Inception backbone vs other backbones (the statistical analysis above the figures is performed on results with IWCV).}
\label{fig:pairwise-inception-vs-original}
\end{figure}

The previous section underlined the importance of a good backbone to extract temporal features relevant to the classification task. 
In this paragraph, we investigate how impactful the choice of the backbone is to the domain adaptation problem.
%if the two most successful backbones, namely 1D-CNN and Inception are statistically equivalent.
Indeed, while Fig.~\ref{fig:all_algo_hyp} indicates slightly better ranks for Inception-based approach, we see in Fig.~\ref{fig:top1-da-method_paper} that changing the Raincoat’s backbone into Inception does not have a significant impact on the actual target accuracy.
% This conclusion holds for CoTMix and CoDATS as shown in Appendix~\ref{app:furth_analysis_backbones}.

%\noindent 
%\begin{minipage}{\textwidth}
%\vspace{.3cm}
%\begin{minipage}{0.47\textwidth}
%\centering
%\includegraphics[width=\linewidth]{fig/Other Backbones-vs-Inception.pdf}
%\captionof{figure}{Pairwise comparison of Inception backbone vs other backbones.}
%\label{fig:pairwise-backbone}
%\end{minipage}
%\hfill
%\begin{minipage}{0.47\textwidth}
%    \centering
%    \small
%    \begin{tabular}{lccc}
%        \toprule
%        & Wins & Ties & Loss  \\
%        \midrule
%        Machinery & \textbf{21} & 0 & 15  \\
%        Medical & \textbf{27} & 3 & 15 \\
%        Motion & 23 & 0 & \textbf{43} \\
%        Remote sensing & \textbf{11} & 0 & 4 \\ 
%        %\midrule
%        %\textit{All themes} & \textit{82} &\textit{3} & \textit{77}  \\
%        \bottomrule
%\end{tabular}
%\captionof{table}{Pairwise accuracy comparison between original backbone against Inception backbone for the 4 different dataset themes, all tuned with IWCV.}
%\label{tab:theme}
%\end{minipage}
%\vspace{.3cm}
%\end{minipage}

\paragraph{Inception and CNN-based backbones are not statistically different}
Fig.~\ref{fig:top1-da-method_paper} compares the two top approaches with IWCV, namely Raincoat and InceptionRain. 
It suggests that a change in backbone does not lead to statistically better results with a $p-value = 0.276$, providing evidence that the DA technique, rather than its backbone, is the main reason behind the performance of InceptionRain in this benchmark.   
To further validate this across all backbones, Fig.~\ref{fig:all-da-method} displays the pairwise accuracy comparison between CoDATS, Raincoat and CoTMix against the three same methods based on the Inception backbone. 
The statistical comparison indicates a small difference with a $p-value$ above $0.8$, thus suggesting that given the current benchmark, backbones other than VRNN do not have a significant impact and the main difference stems from the UDA technique itself.
Note however that there is a couple of datasets (miniTime and sportsActivity) where InceptionMix always predicts the same classes thus has a very low accuracy, suggesting also a collapse during training. As this is also true for the Target Risk, the reason does not seem to be the hyperparameter search but rather an unexpected mismatch between those two specifics datasets, CoTMix algorithm and the Inception backbone.

%\paragraph{Still, 1D-CNN performs better on motion datasets and Inception on other themes}
%Figs.~\ref{fig:theme} 
% Table~\ref{tab:theme} provides a deeper analysis on the four different themes of dataset.%, as well as Fig.~\ref{fig:theme} in Appendix~\ref{app:furth_analysis_backbones}.
% Note that the number of data points being sometimes quite small, the confidence on the statistical analysis is small. Therefore, no statistical difference has been highlighted with Wilcoxon signed-rank tests.
% The following notable observation remains. %First, 
% The original backbones outperform Inception in datasets focused on motion, whike  Inception outperforms the original backbones in all other themes: machinery, medical and remote sensing.
% This may be due to the majority of established TSC UDA datasets (three out of five) being related to motion, suggesting that these models were designed specifically for such datasets.

\paragraph{CNN-based Better for Motion Datasets, Inception Best for Other Themes}
%Figs.~\ref{fig:theme} 
Table~\ref{tab:theme} provides a deeper analysis on the four different themes of dataset however, it should be noted that the number of some experiments per domain is relatively small, thus limiting the confidence in the statistical analysis. As a result, no significant differences were identified using the Wilcoxon signed-rank tests. Nevertheless, the results indicate that the original backbones outperform Inception on motion datasets, while Inception shows better performance on datasets related to other themes, such as machinery, medical, and remote sensing. This discrepancy may be explained by the fact that the majority of established TSC UDA datasets (three out of five in our benchmark) are related to motion, suggesting that the original backbones were specifically designed for these datasets.

\begin{table}[!ht]
    \small
    \centering
    \caption{Pairwise accuracy comparison between original backbone against Inception backbone for the 4 different dataset themes, all tuned with IWCV. The $p-values$ are computed from Wilcoxon signed-rank tests.}
    \label{tab:theme}
    %\small
    \begin{tabular}{lcccc}
        \toprule
        & Inception wins & Ties & Original wins & $p-value$ \\
        \midrule
        Machinery & \textbf{21} & 0 & 15 & 0.388 \\
        Medical & \textbf{27} & 3 & 15 & 0.079\\
        Motion & 23 & 0 & \textbf{43} & 0.079 \\
        Remote sensing & \textbf{11} & 0 & 4 & 0.252\\ 
        %\midrule
        %\textit{All themes} & \textit{82} &\textit{3} & \textit{77}  \\
        \bottomrule
\end{tabular}
\end{table}

%\begin{figure}[H]
%\centering
%\begin{subfigure}{0.49\textwidth}
%    \includegraphics[width=\textwidth]{fig/backbones_by_themes/machinery-Other Backbones-vs-Inception.pdf}
%    \caption{Machinery theme}
%    \label{fig:themebackbone1}
%\end{subfigure}
%\hfill
%    \includegraphics[width=\linewidth]{fig/backbones_by_themes/medical-Other Backbones-vs-Inception.pdf}
%    \caption{Medical theme}
%    \label{fig:themebackbone2}
%\end{subfigure}
%
%\begin{subfigure}{0.49\textwidth}
%    \includegraphics[width=\linewidth]{fig/backbones_by_themes/motion-Other Backbones-vs-Inception.pdf}
%    \caption{Motion theme}
%    \label{fig:themebackbone3}
%\end{subfigure}
%\hfill
%\begin{subfigure}{0.49\textwidth}
%    \includegraphics[width=\linewidth]{fig/backbones_by_themes/remote_sensing-Other Backbones-vs-Inception.pdf}
%    \caption{Remote sensing theme}
%    \label{fig:themebackbone4}
%\end{subfigure}
%     \caption{
%        Pairwise accuracy comparison between original backbone against Inception backbone for the 4 different dataset themes, all tuned with IWCV.
%        }
%        \label{fig:theme}
%\end{figure}

\subsection{Comparison of hyperparameter tuning methods}\label{section:hyp_method}

%We now delve further into the analysis of hyperparameter tuning approaches. 

\begin{figure}[ht!]
     \centering
     \begin{subfigure}{0.49\linewidth}
         \centering
         \includegraphics[width=\linewidth]{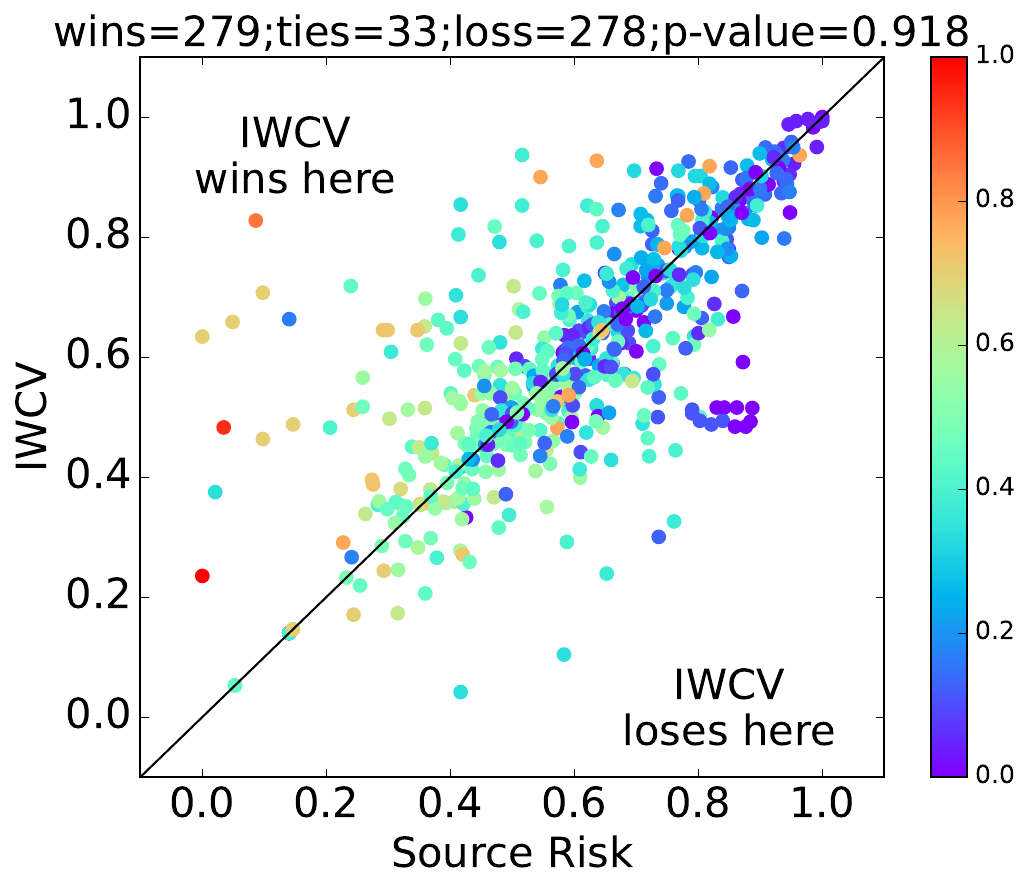}
         \caption{IWCV vs Source Risk}
         \label{fig:reweight-vs-source}
     \end{subfigure}
     \begin{subfigure}{0.49\linewidth}
         \centering
         \includegraphics[width=\linewidth]{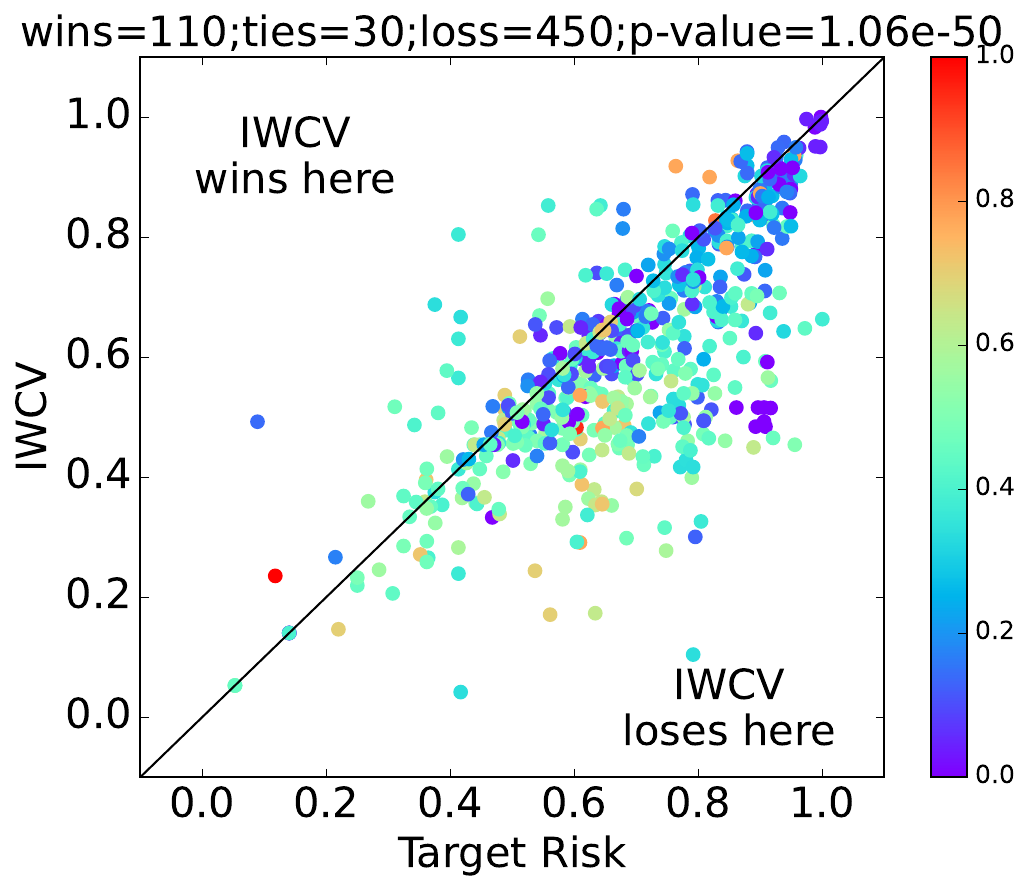}
         \caption{IWCV vs Target Risk}
         \label{fig:reweight-vs-target}
     \end{subfigure}
        \caption{Pairwise accuracy comparison of three hyperparameter tuning methods. Colors correspond to the shift proxy $SP$, with purple for low shifts and red for large ones. }
        \label{fig:pairwise-hparams}
\end{figure}

%First, %the summary depicted in Table~\ref{tab:acc_summary_dts} illustrates the average accuracy per dataset. This summary considers 
In this section we further investigate the hyperparameter tuning approaches. Table~\ref{tab:acc_summary_dts} presents the accuracy per dataset averaged over all classifiers evaluated using the three model selection methods: Source Risk, IWCV and Target Risk. Fig.~\ref{fig:pairwise-hparams} displays the pairwise accuracy comparison with IWCV against Source Risk and Target Risk, with the colormap representing the estimated shift $SP$ (see \eqref{eq:shift-proxy}).

\begin{table}[ht]
\small
\centering
\caption{Accuracy results summary table of average accuracy per dataset valuated the 3 model selection methods: Source Risk, IWCV and Target Risk. %The latter is presented in italic to highlight the fact that it is not available in practice.
}
\label{tab:acc_summary_dts}
\begin{tabular}{l | c c | c}
\toprule
         dataets &            IWCV &     Source Risk &     {Target Risk} \\
\midrule
         bearing & 0.83 $\pm$ 0.03 & \textbf{0.84 $\pm$ 0.02} & 0.86 $\pm$ 0.04 \\
             ecg & 0.62 $\pm$ 0.02 & \textbf{0.63 $\pm$ 0.02} & 0.64 $\pm$ 0.02 \\
       equations & \textbf{0.52 $\pm$ 0.02} & 0.51 $\pm$ 0.02 & 0.57 $\pm$ 0.02 \\
            ford & 0.57 $\pm$ 0.05 & \textbf{0.81 $\pm$ 0.02} & 0.82 $\pm$ 0.02 \\
             har &  \textbf{0.80 $\pm$ 0.05} & 0.73 $\pm$ 0.04 & 0.85 $\pm$ 0.04 \\
            hhar &\textbf{ 0.65 $\pm$ 0.04} & \textbf{0.65 $\pm$ 0.03} & 0.76 $\pm$ 0.03 \\
             mfd & 0.65 $\pm$ 0.09 & \textbf{0.66 $\pm$ 0.09} & 0.75 $\pm$ 0.07 \\
        miniTime & 0.67 $\pm$ 0.05 & \textbf{0.69 $\pm$ 0.04} & 0.72 $\pm$ 0.04 \\
          muscle & \textbf{0.55 $\pm$ 0.03} & 0.52 $\pm$ 0.04 & 0.65 $\pm$ 0.05 \\
           sleep & 0.67 $\pm$ 0.04 &\textbf{ 0.68 $\pm$ 0.04} & 0.75 $\pm$ 0.02 \\
          sports & \textbf{0.54 $\pm$ 0.06} & 0.52 $\pm$ 0.08 & 0.64 $\pm$ 0.08 \\
           wisdm & \textbf{0.54 $\pm$ 0.06 }& 0.46 $\pm$ 0.07 & 0.62 $\pm$ 0.05 \\
\bottomrule
\end{tabular}
\end{table}

Fig.~\ref{fig:reweight-vs-target} supports that having access to target labels is the most effective method for hyperparameter selection using Target risk with a clear advantages over IWCV. This is considered an upper bound due to the practical difficulty of obtaining such labels.
However, as display on Fig.~\ref{fig:reweight-vs-source}, the difference between the IWCV and Source Risk methods is not substantial, as indicated by their similar performances and a Wilcoxon signed-rank test, which clearly reject the hypothesis of statistical difference between the two approaches, yielding a $p$-value close to 1. 
Interestingly, a closer examination the color scale of Fig.~\ref{fig:reweight-vs-source} shows that the larger the shift between source and target (red and yellow), the better IWCV is compare to Source Risk.
Similarly, when the shift between source and target is getting smaller (purple), IWCV can even be harmful, losing to Source Risk.

%Both visualizations further supports that having access to target labels leads to the most effective for hyperparameter selection with Target risk, which we consider here as an upper bound due to the difficulty of getting such labels in practical scenarios. The difference between the IWCV and Source Risk methods is again not substantial, indicated by similar performances and by a Wilcoxon signed-rank test rejecting the hypothesis that both approaches are statistically different with a $p-value$ close to $1$.

%However, looking more closely at Fig.~\ref{fig:pairwise-hparams}, the colors seem to indicate that the larger the shift between source and target, the larger the difference between IWCV and Source Risk. 
%
%Table~\ref{tab:acc_summary_dts} also highlights the difficulties or the challenges of the different datasets, as evidenced by lower performance even with Target risk. Notably, the most challenging ones include OnHWeq, wisdm, sports activities, and sleep stages. Among these datasets, three present high shift between source and target (OnHWeq, wisdm, and sports activities), and two stand out as particularly imbalanced (Wisdm and sleep stages), and further complicating the classification task. OnHWeq also presents difficulty due to its high number of classes (15 classes).

Additionally, Table~\ref{tab:acc_summary_dts} also highlights the difficulties or the challenges of the different datasets. Notably, the most challenging ones include equations, egc, wisdm, sports, and muscle, where average performance with Target risk are lower than 70\%. 
Among these datasets, four present high shift between source and target (equations, wisdm, muscle and sports),
% and two stand out as particularly imbalanced (ecg and wisdm),
which further complicating the classification task.
Both datasets equations and sports present extra difficulty due to their high number of classes (15 and 19 classes respectively). 
Therefore, we propose to further investigate the impact of the shift and other dataset characteristics
%in the next paragraph, while having a deeper look into other dataset characteristics in the following section,
in order to better understand the performance of UDA algorithms and hyperparameter tuning approaches.

%Table~\ref{tab:acc_summary_clf} reaffirms the observations and conclusions derived from the critical diagrams presented in this benchmark, concerning the ranking of algorithms and the performance of the backbones.

% \begin{figure}[ht!]
%      \centering
%      \begin{subfigure}{0.49\linewidth}
%          \centering
%          \includegraphics[width=\linewidth]{fig/TransportedSource_vs_Reweight_5_bis.pdf}
%          \caption{IWCV vs Source Risk}
%          \label{fig:reweight-vs-source}
%      \end{subfigure}
%      \begin{subfigure}{0.49\linewidth}
%          \centering
%          \includegraphics[width=\linewidth]{fig/TransportedTarget_vs_Reweight_5_bis.pdf}
%          \caption{IWCV vs Target Risk}
%          \label{fig:reweight-vs-target}
%      \end{subfigure}
%         \caption{Pairwise accuracy comparison of three hyperparameter tuning methods. Colors correspond to the shift proxy $SP$, with purple for low shifts and red for large ones. }
%         \label{fig:pairwise-hparams}
% \end{figure}

%\paragraph{IWCV improves performance for larger shifts, while Source risk is better for low shifts}

\begin{figure}
    \centering
    \begin{subfigure}{.32\linewidth}
    \includegraphics[width=\linewidth]{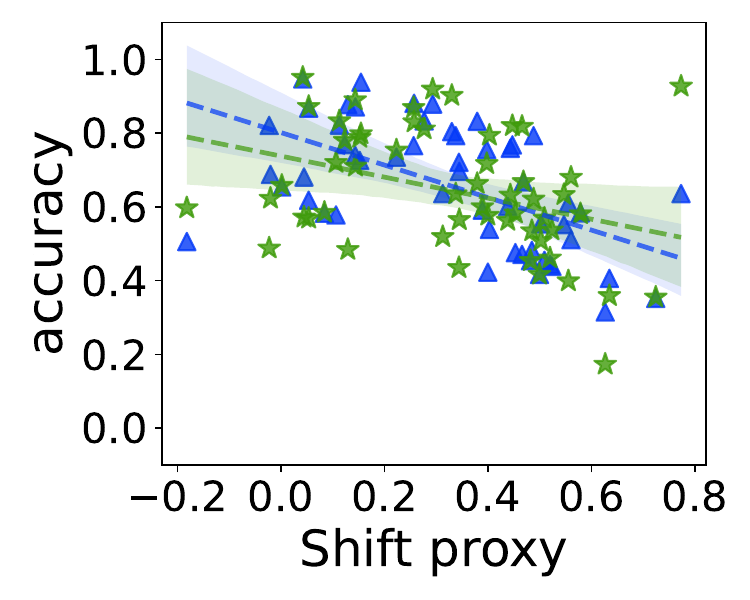}
    \caption{CoDATS}
    \label{fig:shift-codats}
    \end{subfigure}
    \begin{subfigure}{.32\linewidth}
    \includegraphics[width=\linewidth]{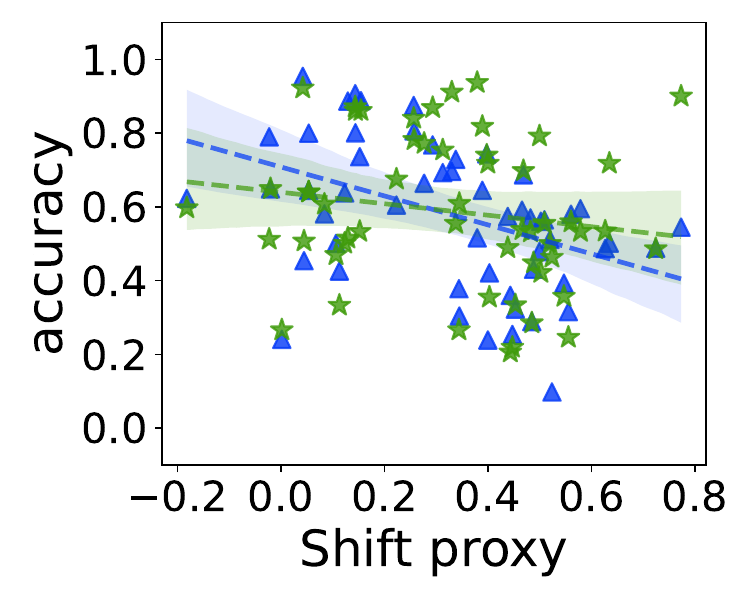}
    \caption{CoTMix}
    \label{fig:shift-cotmix}
    \end{subfigure}
    \begin{subfigure}{.32\linewidth}
    \includegraphics[width=\linewidth]{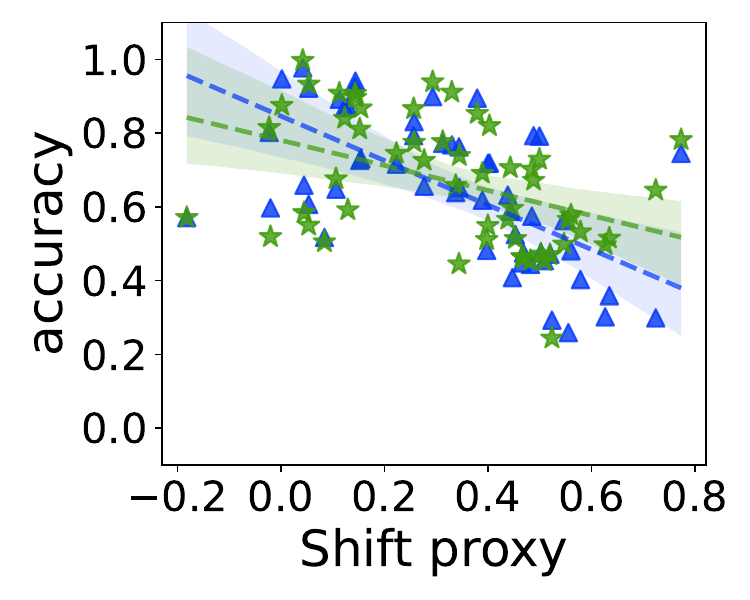}
    \caption{Raincoat}
    \label{fig:shift-raincot}
    \end{subfigure}
    \begin{subfigure}{.32\linewidth}
    \includegraphics[width=\linewidth]{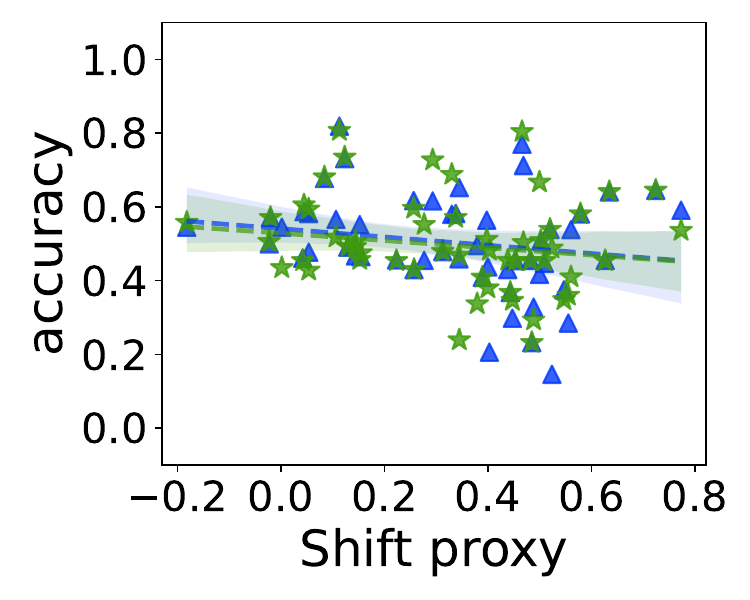}
    \caption{VRADA}
    \label{fig:shift-vrada}
    \end{subfigure}
    \begin{subfigure}{.32\linewidth}
    \includegraphics[width=\linewidth]{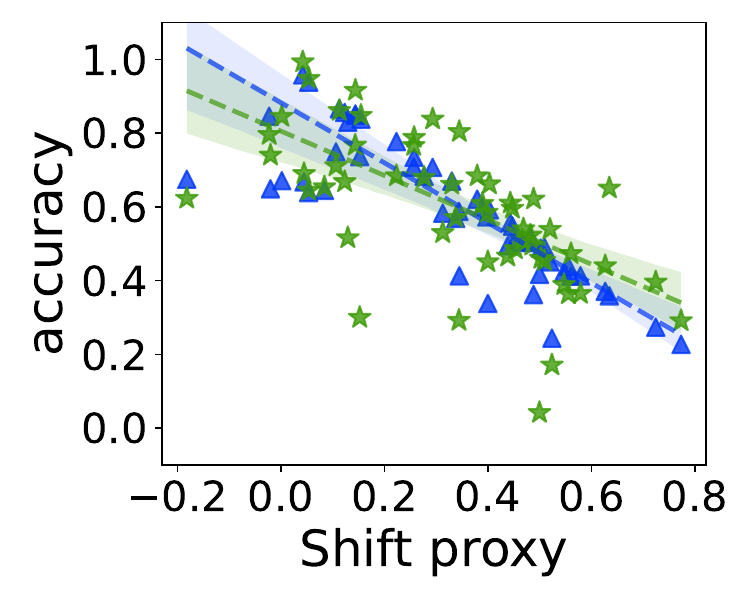}
    \caption{Inception (no DA)}
    \label{fig:shift-source}
    \end{subfigure}
    \caption{Impact of the estimated shift on the performance of UDA algorithms tuned with either IWCV (\textcolor{vertFonce}{\small{$\bigstar$}}) or Source Risk (\textcolor{blue}{$\blacktriangle$}). The dashed lines display a linear regression.}
    \label{fig:perf-vs-shift}
\end{figure}

\paragraph{IWCV Outperforms with Large Shifts, Source Risk Favored for Small ones}
Fig.~\ref{fig:perf-vs-shift} shows how the shift, estimated with $SP$, affects the performance of UDA algorithms tuned with either IWCV or Source Risk. 
%
%\begin{figure}[ht]
%    \centering
%    \includegraphics[width=\linewidth]{fig/meta_analysis/accuracy_vs_shift_all_both.pdf}
%    \caption{Impact of the estimated shift on the performance of algorithms tuned with either IWCV (\textcolor{vertFonce}{\small{$\bigstar$}}) or Source Risk (\textcolor{blue}{$\blacktriangle$}). The lines display a linear regression. %\textcolor{red}{Hassan: I think we can remove the legends from the figure and keep it simply in the caption. We can use: (red dot vs blue star) to differentiate between the two easily. We can also reduce the number of ticks in the y axis similar to what we did for the x axis. Other than that I think this is converging!!! we can use same comment for the figure 8}
%    % we keep the colors as is as they are the same we used in other figures: IWCV in green and Source risk in blue
%    }
%    \label{fig:perf-vs-shift}
%\end{figure}
%
%Fig.~\ref{fig:pairwise-hparams} displays the pairwise accuracy comparison with IWCV against Source Risk and Target Risk, with the colormap representing the estimated shift SP~(\ref{eq:shift-proxy}).
%Fig.~\ref{fig:reweight-vs-source} indicates that IWCV and Source Risk are not significantly different with a $p-value$ close to $1$, but the colors indicate that the larger the shift between source and target, the larger the difference between IWCV and Source Risk. 
%Looking closely at the relationship between shift level and hyperparameter tuning, 
From this figure, and setting aside VRADA which has very bad performance overall,
%both approaches lead to similar performance,
we can see a tendency, with all methods, of IWCV surpassing Source risk on datasets with large shift between source and target, while Source Risk tends to perform better for datasets with low shifts.
The latter can be explained by the similarity of the Source Risk and Target Risk in the presence of small domain shifts.
On the contrary, the estimation of the target risk with IWCV is theoretically more accurate regardless of the shift but the challenging density estimation adversely affects performance at lower shift values.
Note that similar behavior can be observed with those method with Inception backbone.
The similitude in performance of both tuning approaches %Figs.~\ref{fig:all_algo_hyp} and \ref{fig:reweight-vs-source} 
in previous tables and figures can be attributed to the variety of shifts in our benchmark datasets.

Nevertheless, Fig.~\ref{fig:reweight-vs-target} shows that IWCV (and thus also Source Risk) performs  significantly worse than Target Risk, highlighting the need for further research on hyperparameter tuning methods to reduce this large gap.

\subsection{Impact of dataset characteristics}
\begin{figure}[!ht]
    \centering
    \begin{subfigure}{.25\linewidth}
    \includegraphics[width=.95\linewidth]{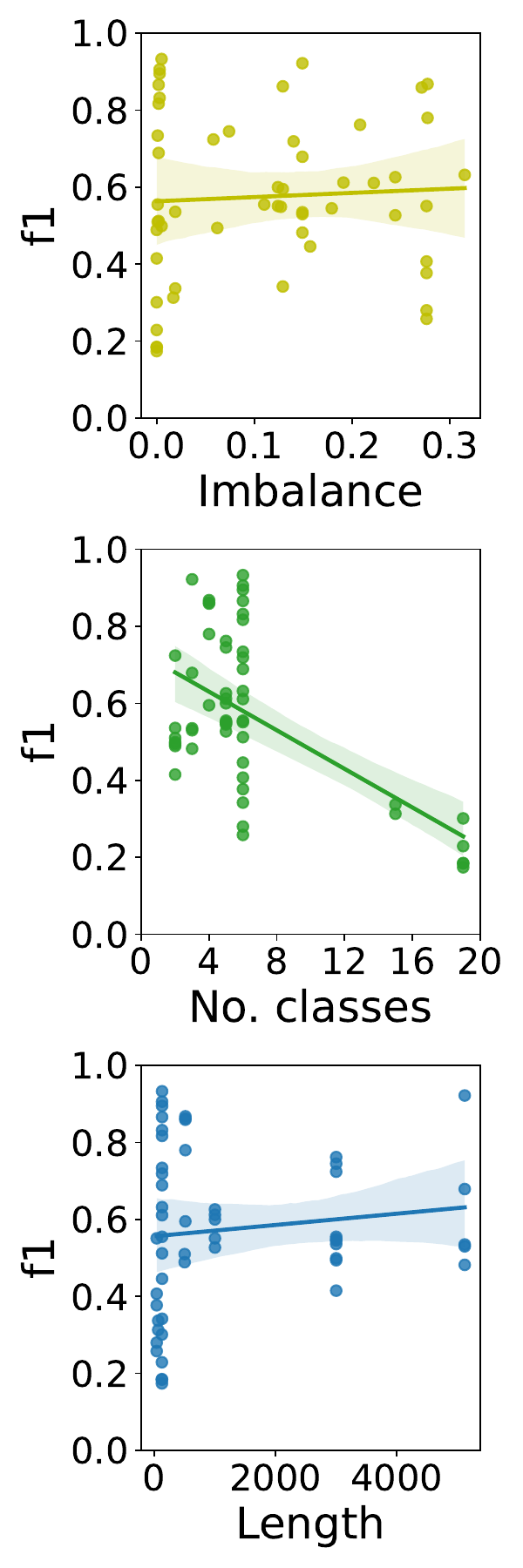}
    \caption{CoTMix}
    \label{fig:meta-analysis-cotmix}
    \end{subfigure}
    \begin{subfigure}{.25\linewidth}
    \includegraphics[width=.95\linewidth]{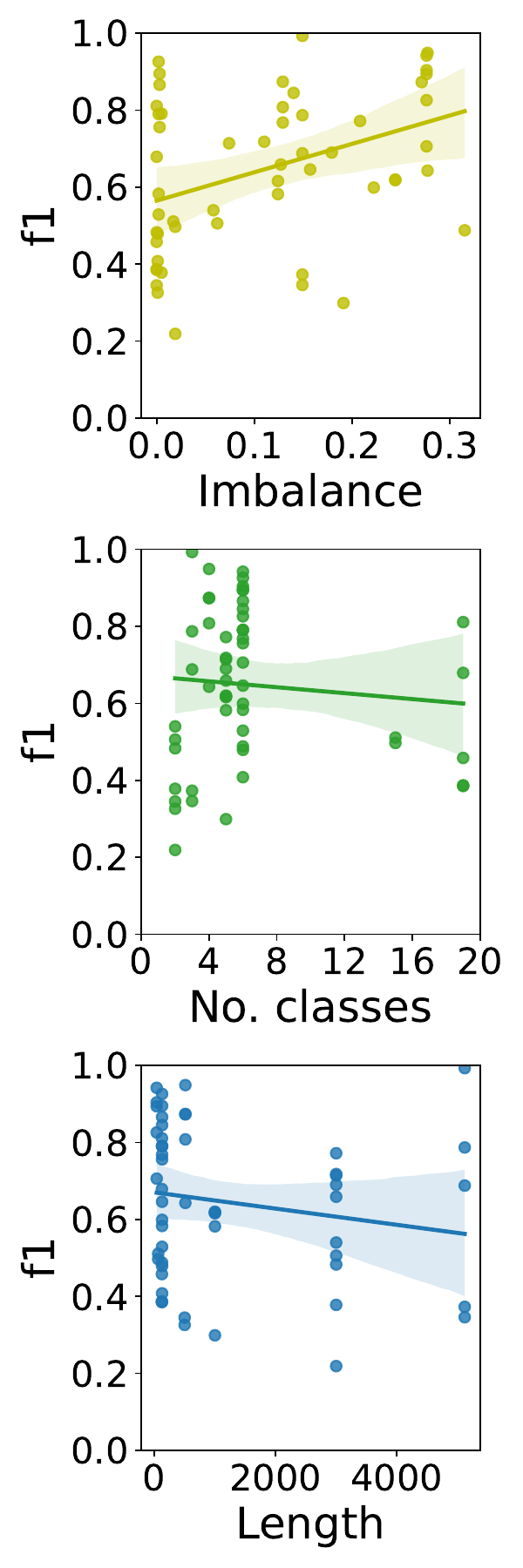}
    \caption{InceptionRain}
    \label{fig:meta-analysis-inceptionrain}
    \end{subfigure}
    \begin{subfigure}{.25\linewidth}
    \includegraphics[width=.95\linewidth]{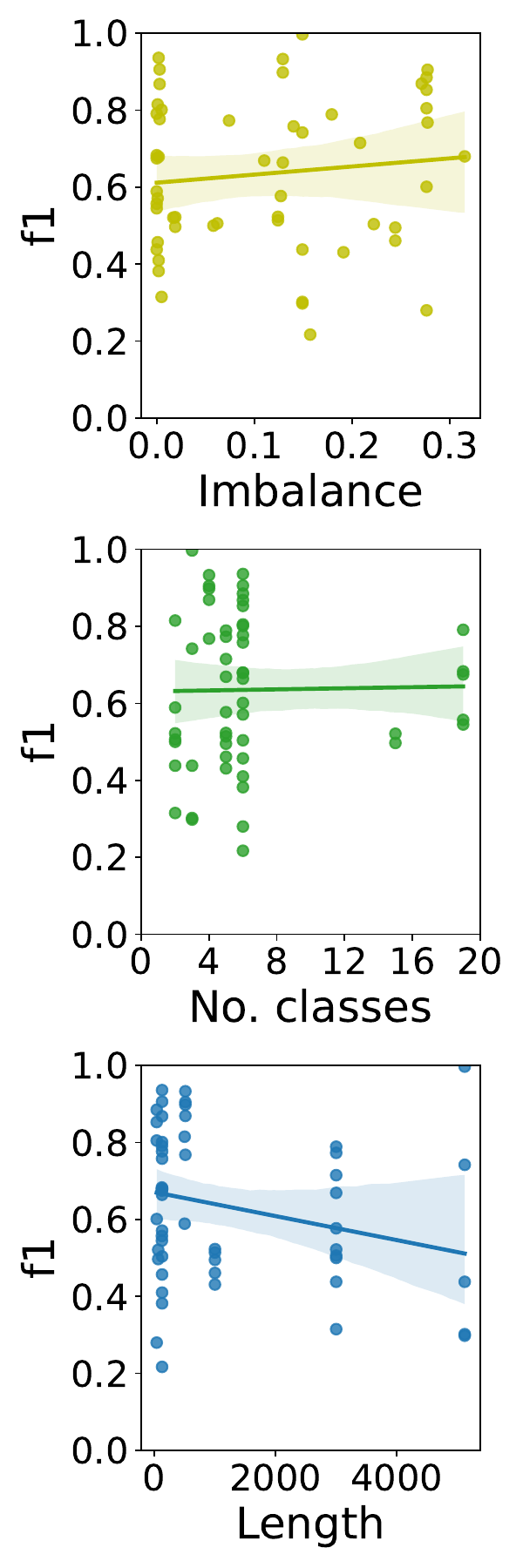}
    \caption{Raincoat}
    \label{fig:meta-analysis-raincoat}
    \end{subfigure}
    \caption{Impact of dataset characteristics on performance (measured in F1-score) of UDA algorithms tuned with IWCV.
    %\todo{[todo: add legend everywhere + other examples]}
    }
    \label{fig:meta-analysis}
\end{figure}

In this final section of results, we study and examine the high-level impact of dataset characteristics on the performance (in F1-score) of UDA algorithms. Fig.~\ref{fig:meta-analysis} displays the performance with respect to the several dataset characteristics; the imbalance of the datasets, %the number of time series in the dataset (No. instances), 
the length of the time series and the number of classes (which is highly correlated to the number of channels in our benchmark datasets). %Note that Fig.~\ref{fig:meta-analysis} highlights the main behaviors

% \begin{figure}[!ht]
%     \centering
%     \begin{subfigure}{\linewidth}
%     \includegraphics[width=\linewidth]{fig/meta_analysis/CoDATS_f1_vs_all_metadata.pdf}
%     \caption{CoDATS}
%     \label{fig:meta-analysis-codats}
%     \end{subfigure}
%     \begin{subfigure}{\linewidth}
%     \includegraphics[width=\linewidth]{fig/meta_analysis/CoTMix_f1_vs_all_metadata.pdf}
%     \caption{CoTMix}
%     \label{fig:meta-analysis-cotmix}
%     \end{subfigure}
%     \begin{subfigure}{\linewidth}
%     \includegraphics[width=\linewidth]{fig/meta_analysis/Raincoat_f1_vs_all_metadata.pdf}
%     \caption{Raincoat}
%     \label{fig:meta-analysis-raincoat}
%     \end{subfigure}
%     \begin{subfigure}{\linewidth}
%     \includegraphics[width=\linewidth]{fig/meta_analysis/VRADA_f1_vs_all_metadata.pdf}
%     \caption{VRADA}
%     \label{fig:meta-analysis-vrada}
%     \end{subfigure}
%     \caption{Impact of dataset characteristics on performance (measured in F1-score) of UDA algorithms tuned with IWCV.
%     %\todo{[todo: add legend everywhere + other examples]}
%     }
%     \label{fig:meta-analysis}
% \end{figure}

\paragraph{Most UDA algorithms are stable across different data characteristics, only CoTMix is highly impacted by the number of classes.}
The most striking impact can be observed from Fig.~\ref{fig:meta-analysis-cotmix}, where it is very clear that CoTMix performance decreased significantly with an increased number of classes, %potentially explaining the approach's poor performance on some datasets and its overall ranking.
which is consistent with our hypothesis that the temporal mixup strategy does not separate well the classes. Interestingly, CoTMix (and also InceptionMix) seems to result in slightly better performance for longer time series contrary to the other methods.
%InceptionMix displays a similar behavior.
%For the other approaches and dataset characteristics, the impact is much less clear. 
%Another noteworthy aspect concerns the impact of the dataset imbalance for InceptionRain which is also true for InceptionDANN and InceptionCDAN. These methods exhibit a rise in performance for highly imbalanced datasets, which seems counterintuitive but can be 
Another noteworthy aspect concerns the rise or stability of the performance for imbalanced datasets, which seems counterintuitive but can be explained by overall simpler tasks for those datasets.
%Finally, the remaining UDA approaches, namely Raincoat, CoDATS and VRADA, demonstrate stability across dataset characteristics.
%, contributing to their strong performances.
%Similarly, %We can observe from Fig.~\ref{fig:meta-analysis-vrada} that 
%VRADA consistently underperforms across all dataset conditions, although it seems to improve slightly with the number of data instances.
%Despite the noisy data, we observe no decline in performance from the Inception backbone under any condition, further endorsing this choice of backbone.
%On  In contrast, the other top-performing approaches, such as InceptionRain, demonstrate greater stability across dataset characteristics, contributing to their strong performances.

% \begin{figure}[!ht]
%     \centering
%     %\begin{subfigure}{0.7\linewidth}
%     %\includegraphics[width=\linewidth]{fig/meta_analysis/Source only_vs_VRADA_accuracy_vs_metadata_Main.pdf}
%     %\caption{Inception (no DA) and VRADA}
%     %\label{fig:meta-analysis-cotmix-raincoata}
%     %\end{subfigure}
%    \begin{subfigure}{0.7\linewidth}
%    \includegraphics[width=\linewidth]{fig/meta_analysis/CoTMix_vs_InceptionRain_accuracy_vs_metadata_Main.pdf}
%    \caption{CoTMix and InceptionRain}
%    \label{fig:meta-analysis-cotmix-raincoatb}
%    \end{subfigure}
%    \caption{Impact of number of classes and class imbalance on performance of selected UDA algorithms. \todo{[todo: add legend everywhere + other examples]}}
%    \label{fig:meta-analysis-cotmix-raincoat}
% \end{figure}

\section{Conclusion}

This study presents a comprehensive benchmark evaluation of contemporary algorithms for deep unsupervised domain adaptation (UDA) in the context of time series classification. 
Additionally, we adapted datasets to domain adaptation to establish a strong baseline for performance evaluation, facilitating a clearer understanding of the research landscape.
This benchmark enables a fair comparison among algorithms by employing various hyperparameter tuning methods while maintaining a consistent time budget for the tuning process.

The main conclusions of our study are as follows. The top performing UDA approaches were found to be Raincoat, which is tailored for time series, and those based on domain adversarial techniques.
The backbone choice seems to be less important than the UDA strategy, yet applying the Inception backbone to Raincoat performs slightly better.
For tuning hyperparameters, IWCV and Source risk were found to perform similarly, although IWCV leads to improved performance for higher shifts, while Source Risk is robust for lower shifts. 
Still, both tuning methods perform significantly worse than the Target Risk, suggesting the potential benefit for hyperparameter tuning methods tailored for time series data.
Finally, we provided useful insights to how UDA approaches are affected by data characteristics, so that practitioners can select the UDA approach that aligns with their requirements.

% \begin{appendices}

% \include{appendix}

% \end{appendices}


\bibliography{sn-bibliography}% common bib file

%\appendix
%\input{appendix}

\end{document}

%% file: mathinput.tex
\usepackage{amsmath,amsfonts,bm}

\def\RR{\mathbb{R}}

\def\eqref#1{equation~\ref{#1}}
\def\1{\bm{1}}

\def\rvy{{\mathbf{y}}}

\def\rmX{{\mathbf{X}}}

\DeclareMathAlphabet{\mathsfit}{\encodingdefault}{\sfdefault}{m}{sl}
\SetMathAlphabet{\mathsfit}{bold}{\encodingdefault}{\sfdefault}{bx}{n}

\newcommand{\Ls}{\mathcal{L}}